\pdfoutput=1
\documentclass[conference]{IEEEtran}
\IEEEoverridecommandlockouts
\usepackage{cite}
\usepackage{amsmath,amssymb,amsfonts}
\usepackage{algorithmic}
\usepackage[final]{graphicx}
\usepackage{subfigure}
\usepackage{bm}
\usepackage{booktabs}
\usepackage{caption}
\usepackage{textcomp}
\usepackage{xcolor}
\def\BibTeX{{\rm B\kern-.05em{\sc i\kern-.025em b}\kern-.08em
    T\kern-.1667em\lower.7ex\hbox{E}\kern-.125emX}}
\usepackage{url}    

\begin{document}



\title{ Deep Human-guided Conditional Variational Generative Modeling for Automated Urban Planning}

\author{\IEEEauthorblockN{Dongjie Wang, Kunpeng Liu, Pauline Johnson, Leilei Sun,  Bowen Du, Yanjie Fu\IEEEauthorrefmark{2}}
\IEEEauthorblockA{
\textit{ Department of Computer Science, University of Central Florida, Orlando} \\
\textit{Department of Computer Science, Beihang University, Beijing}\\
\{wangdongjie, kunpengliu, pauline.johnson\}@knights.ucf.edu, \\
\{Yanjie.Fu\}@ucf.edu 
 \{leileisun, dubowen\}@buaa.edu.cn,  \\ 
}
\IEEEcompsocitemizethanks{
	\IEEEcompsocthanksitem \textbullet ~ Yanjie Fu is contact author.
	}

}


\maketitle

\begin{abstract}
Urban planning designs land-use configurations and can benefit building livable, sustainable, safe communities. 
Inspired by image generation, deep urban planning aims to leverage deep learning to generate land-use configurations.
However, urban planning is a complex process. Existing studies usually ignore the need of personalized human guidance in planning, and  spatial hierarchical structure in planning generation. 
 Moreover, the lack of large-scale land-use configuration samples poses a data sparsity challenge.
This paper studies a novel deep human guided urban planning method to jointly solve the above challenges. 
Specifically, we formulate the problem into a deep conditional variational autoencoder based framework. 
In this framework, we exploit the deep encoder-decoder design to generate land-use configurations.
To capture the spatial hierarchy structure of land uses, we enforce the decoder to generate both the coarse-grained layer of functional zones, and the fine-grained layer of POI distributions. 
To integrate human guidance, we allow humans to describe what they need as texts and use these texts as a model condition input. 
To mitigate training data sparsity and improve model robustness, we introduce a variational Gaussian  embedding mechanism. 
It not just allows us to better approximate the embedding space distribution of training data and sample a larger population to overcome sparsity, but also adds more probabilistic randomness into the urban planning generation to improve embedding diversity so as to improve robustness.
Finally, we present extensive experiments to validate the enhanced performances of our method. 

\end{abstract}

\section{Introduction}
Urban planning aims to generate land-use configuration plans, in order to boost commercial activities, enhance public security, foster social interaction, and thus yield livable, sustainable, and viable environments.
Urban planning is critical because poor planning could directly or indirectly lead to high crimes, traffic congestion and accidents, air pollution, depression and anxiety.

Recently, inspired by image generation, deep urban planing, where deep learning is leveraged to automate the generation of land-use configuration, is emerging. 
For example, the study in \cite{wang2020reimagining} proposes a  Generative Adversarial Neural network (GAN) method to automatically generate land-use configuration for a region given surrounding contexts. Although it is appealing to teach a machine to plan city configuration, there are certain limitations with existing literature. 
First, urban planning indeed is a complex process that involves with public policy, social science, engineering, architecture, landscape, and other related field. In real world urban planning practice, civil and regional planning experts participate to guide the entire planning process. Some communities emphasize more accessibility to transportation and groceries, while other communities expect higher ratio in landscape design. Existing literature in deep automated urban planning lacks human machine interaction to provide personalized planning. 
Second, many existing methods directly generate POI distribution at the community level, and lacks the in-depth consideration of spatial hierarchical correlations from functional zones to residential communities. 
Third, city configurations are usually sparse and even imperfect, and, thus, has imposed a high demand for robust learning algorithm in sparse data. 

\begin{figure}[htbp]
\vspace{-0.5cm}
    \centering
    \includegraphics[width=0.4\textwidth]{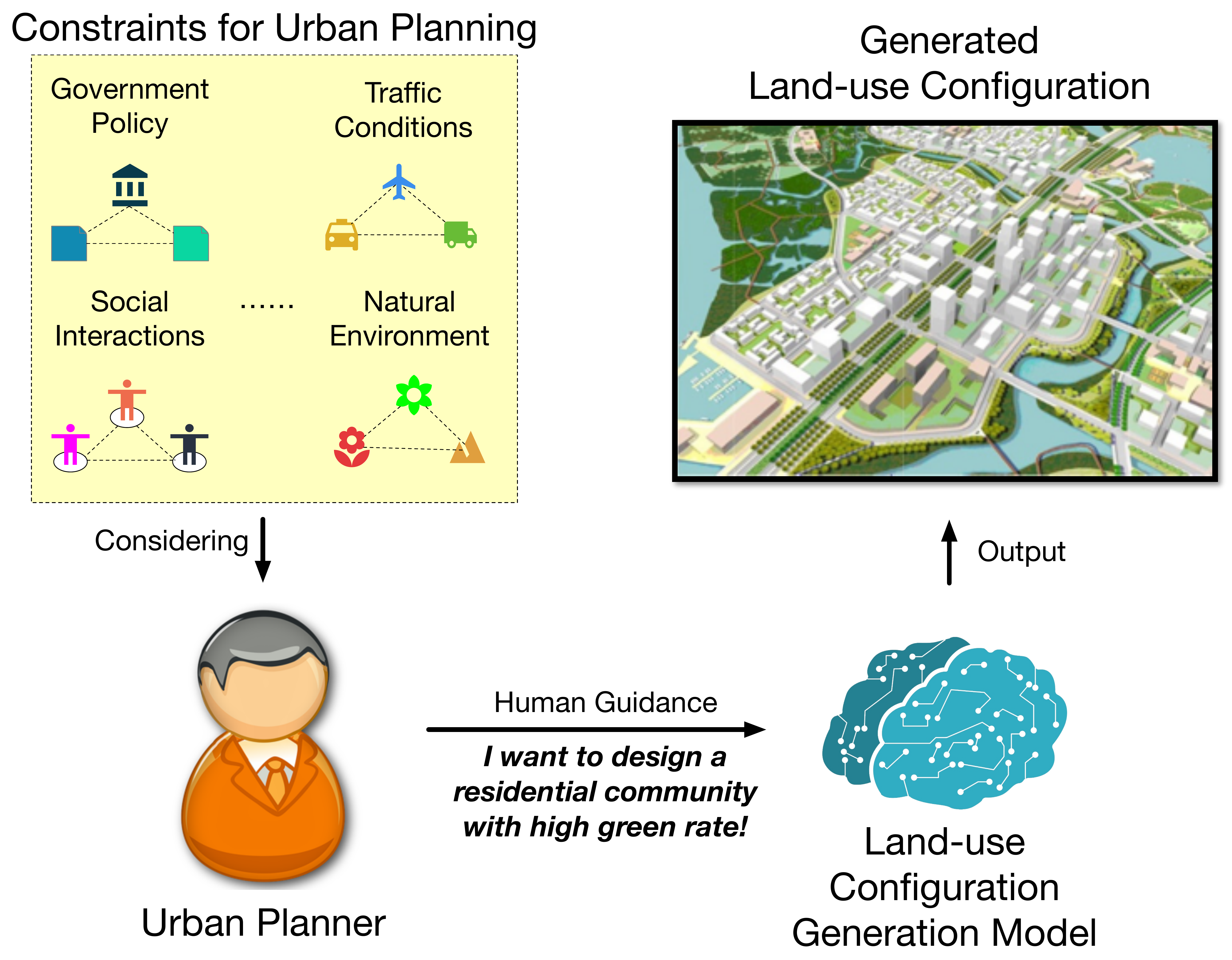}
        \vspace{-0.1cm}
    \caption{Human input textual instructions to guide AI to generate land-use configurations}
    \label{landuse image}
    \vspace{-0.3cm}
\end{figure}

This paper focuses on the problem of deep human-guided urban planning using imperfect urban geographic and human mobility data. 
The basic idea is to formulate the problem into a novel machine learning task of robust generative modeling with spatial hierarchy awareness for sparse data. 
The new machine learning model should have the ability to generate a POI distribution tensor as land-use configuration, to overcome data sparsity, to capture spatial hierarchy structure from functional zones to residential communities in the planning process. 

There are three technical challenges in addressing such machine learning task: 1) \textbf{Challenge 1: human guidance }: in deep urban planing, the guidance from domain experts is needed. So, how can we provide a mechanism to allow humans to interact with and guide deep learning models to generate personalized planning?
2) \textbf{Challenge 2: spatial hierarchy}: 
how can we teach a machine to learn the spatial hierarchy by first planing urban functions at the zone level, then planning POI distribution at the community level? The underlying idea of this strategy is to add structured regularization into the planning generation process.
3) \textbf{Challenge 3: robustness modeling against data sparsity}: community-level configuration data are expensive, imperfect, and sparse. Thus, there is limited quality training data  available. As a result, to overcome such practical data challenge,  we need to answer: how can we equip the planning generative model with robustness against data sparsity?


As a result, we propose a deep conditional variational encoder-decoder framework with awareness of spatial hierarchy to address these challenges jointly. 
The main contributions of our study are summarized as follows: 
1) \textbf{Formulating the automated urban planning into a novel deep encoder-decoder learning task.} The encoder is used to learn the relationship between condition inputs and the corresponding land-use configurations.
The decoder is regarded as a generator to generate the best land-use configurations for planning based on condition inputs. 
2) \textbf{Condition embedding as a bridge to integrate human guidance.}  To allow humans to guide the generation process and provide an interface for human-planner interaction, a human can describe personalized planning needs as texts.
In addition, many factors such as government policy, traffic conditions, etc should be considered for planning reasonable configurations.
Thus, we convert human description texts and the factors in surrounding contexts into embedding respectively, and concatenate the two embeddings as the condition embedding that conditions on both encoder and decoder. 
In this way, human textual inputs are involved in the model learning process.
3) \textbf{Variational additive Gaussian as an augmented tool to fight data sparsity.} To overcome the sparsity issue of land-use configuration data, we make the encoder return a distribution over the latent embedding space instead of a single embedding point. Such distribution can better approximate the data spray pattern in the embedding space, then we can keep drawing samples from the distribution, instead of requiring large-scale quality training configuration data. In addition, this strategy will add randomness in embedding, diversify representations, and improve model robustness. 
4) \textbf{Dual reconstruction to capture spatial hierarchy.} We force the decoder to reconstruct not just the urban function labels of various zones, but also the POI distribution tensor of various communities, in order to capture the zone-community hierarchical correlation.
5) \textbf{Comprehensive experiments to validate the effectiveness of our framework.} 
We evaluated the proposed methods using Beijing's urban geography, human mobility, land-use configuration, POI checkin data. For comparison we implemented a broad range of other algorithms. Results show that the proposed methods consistently outperformed 6 competing methods. We also performed ablation study, sensitivity study, human guided conditional generation study to justify the effectiveness and superiority of our technical insights. 

\section{Definitions and Problem Statement}
\subsection{Definitions}
\subsubsection{Target Area and Surrounding Contexts}
\label{target}
\begin{figure}[htbp]
\vspace{-0.3cm}
    \centering
    \includegraphics[width=0.3\textwidth]{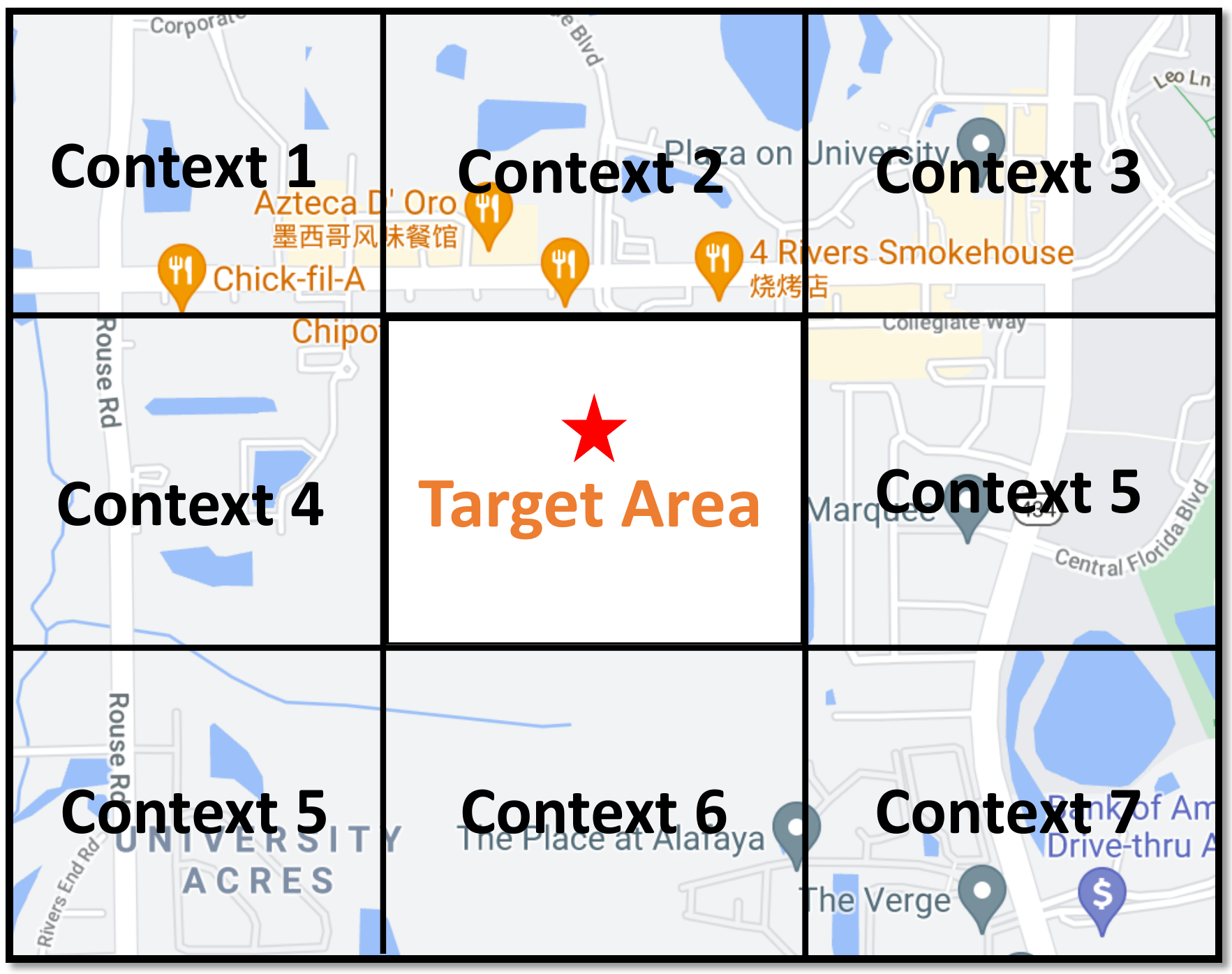}
    \caption{A target area and the surrounding contexts.}
    \label{residential_context}
    \vspace{-0.3cm}
\end{figure}

Target area refers to a geographical square region, in which is empty and waiting for being planned.
Surrounding contexts possess the same square shape as the target area and surround the target area from different directions.
Figure \ref{residential_context} illustrates the spatial relationship between the target area and the surrounding contexts.
There are many socioeconomic activities such as commute, real estate transaction, entertainment, and etc in surrounding contexts.
These socioeconomic characteristics of surrounding contexts affect the design of the land-use configuration of the target area.

\subsubsection{Land-use Configuration}
\label{landuse}
refers to the POI distribution in a geographical region.
To make our model perceive the land-use configuration easily, we adopt the quantitative definition in \cite{wang2020reimagining}.
Figure \ref{landuse image} shows an example of a land-use configuration that has three dimensions: latitude, longitude, and POI category.
To see the details of the configuration further, we pick up the purple layer (\textit{i.e store}) to exhibit.
The latitude and longitude of this layer are divided into $N \times N$ parts. 
The value of each little square represents the number of stores in the corresponding geographical location.
Other layers in the configuration have the same data structure as the purple layer.
Thus, the whole tensor reflects the POI distribution of all POI categories in the geographical region.

\begin{figure}[htbp]
\vspace{-0.3cm}
    \centering
    \includegraphics[width=0.5\textwidth]{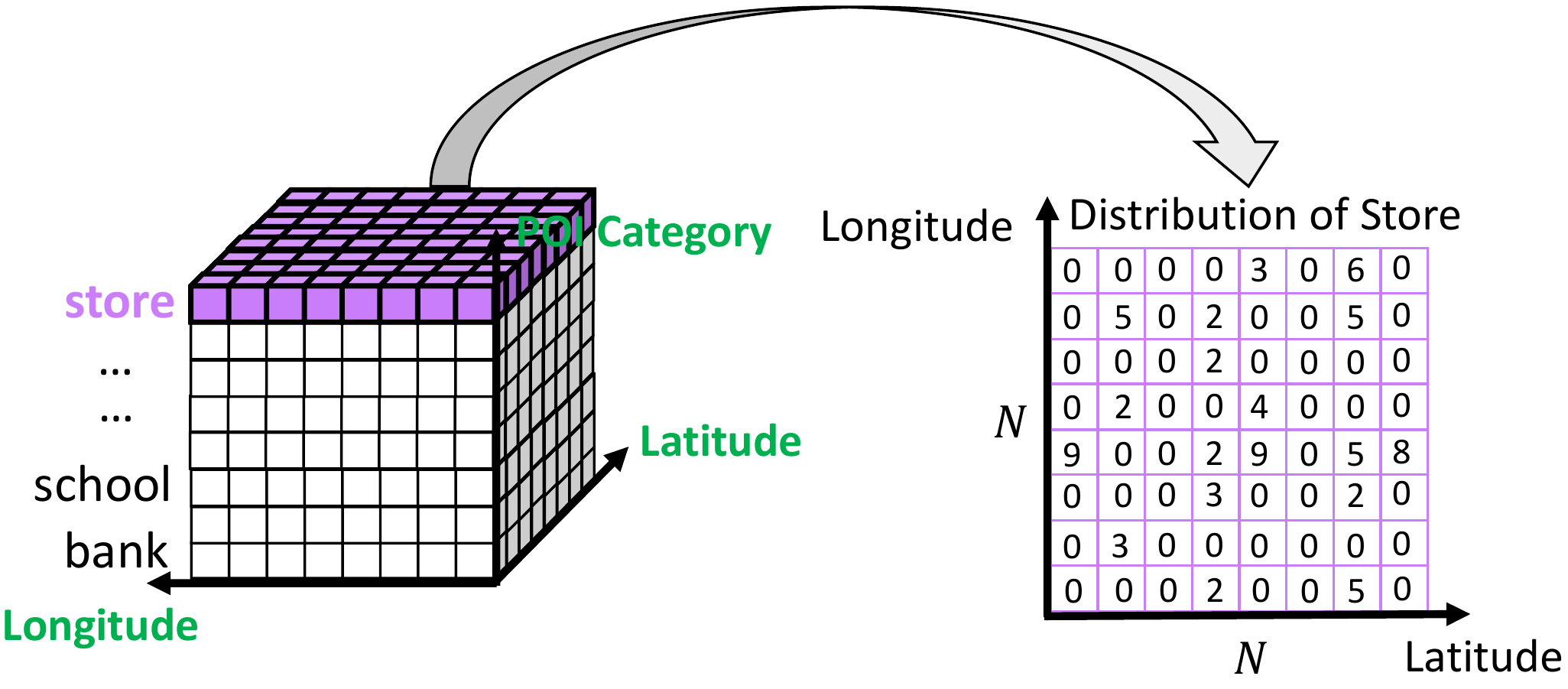}
    \caption{An example of land-use configuration.}
    \label{landuse image}
    \vspace{-0.3cm}
\end{figure}

\subsubsection{Urban Functional Zone} 
\begin{figure}[htbp]
\vspace{-0.3cm}
    \centering
    \includegraphics[width=0.3\textwidth]{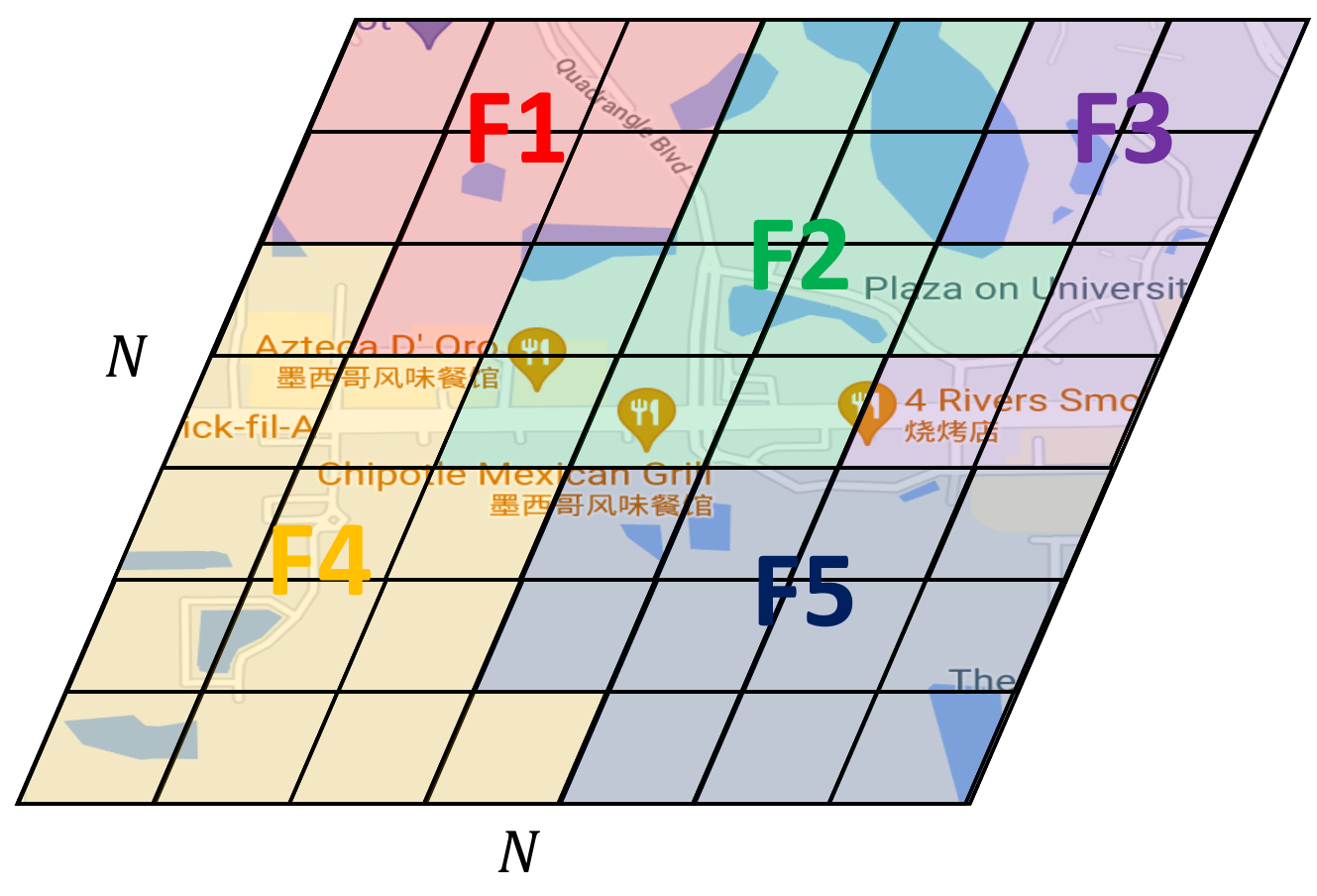}
    \vspace{-0.2cm}
    \caption{An example of urban functional zones.}
    \label{urban_func}
    \vspace{-0.3cm}
\end{figure}

refers to a geographical cluster that owns a specific land-use function, such as industrial zone, educational zone, and etc. 
Compared with land-use configuration, the urban functional zone is more coarse-grained in the urban planning domain. 
Figure \ref{urban_func} show an example of urban functional zones, in which different colors indicate different functional zones.
The same to land-use configuration, we also divide a geographical region into $N \times N$.
Each functional zone consists of certain little squares existing in land-use configurations.
Thus, compared with land-use configuration,  urban functional zones are more coarse-grained and provide a rough sketch for urban planning.

\subsubsection{Human Guidance} 
\label{human}
refers to the humans' requirements for generating a land-use configuration.
Here, these requirements are described by  corresponding texts.
And, we regard the embeddings of these texts as conditional inputs of our land-use configuration generation model.


\begin{figure*}[!t]
    \centering
    \includegraphics[width=1.0\linewidth]{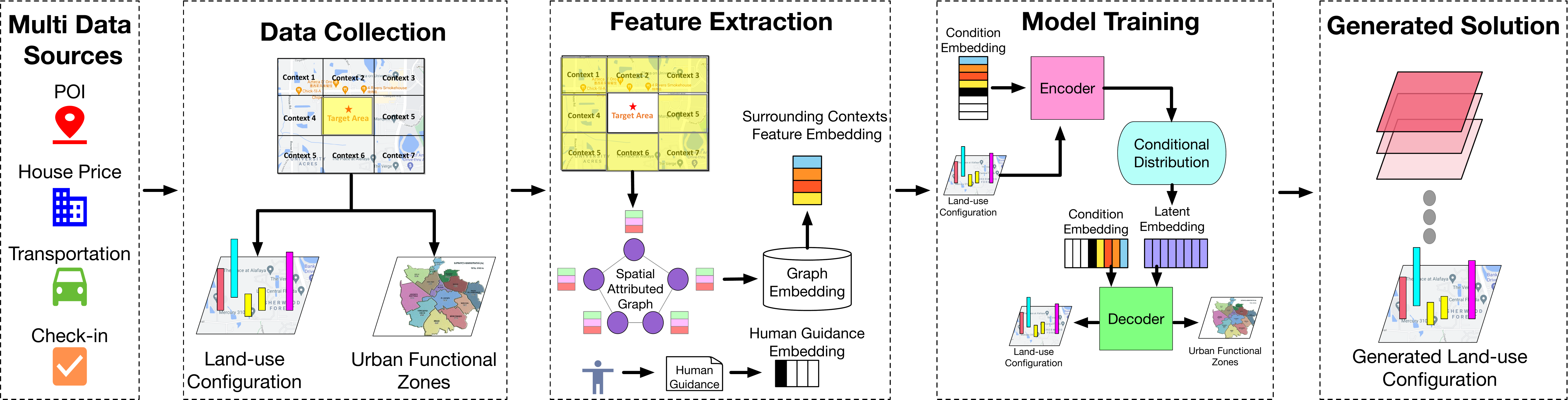}
    \vspace{-0.2cm}
    \captionsetup{justification=centering}
    \caption{Framework Overview.}
    \vspace{-0.6cm}
    \label{fig:framework}
\end{figure*}

\subsection{Problem Statement}
In this paper, we aim to build up an automated land-use configuration generator that can generate configurations via considering the socioeconomic characteristics of surrounding contexts and human guidance.
Formally, given a list of land-use configurations $\mathbf{\tilde{X}} = \{ \mathbf{\hat{X}}^{(1)},\mathbf{\hat{X}}^{(2)},\dots,\mathbf{\hat{X}}^{(K)} \}$, 
socioeconomic features of surrounding contexts $\mathbf{S} = \{ \mathbf{s}^{(1)},\mathbf{s}^{(2)},\dots,\mathbf{s}^{(K)}\}$, 
human guidance $\mathbf{I}=\{ \mathbf{i}^{(1)},\mathbf{i}^{(2)},\dots,\mathbf{i}^{(K)}\}$.
Our purpose is to find the mapping function $f:(\mathbf{S}, \mathbf{I}) \rightarrow \mathbf{\tilde{X}}$ that takes socioeconomic features $\mathbf{S}$ and human guidance $\mathbf{I}$ as input, and outputs the corresponding land-use configurations $\mathbf{\tilde{X}}$.

\section{Proposed Method}
In this section, we introduce the technical details of our proposed framework CLUVAE. 

\subsection{Framework Overview}
Figure \ref{fig:framework} shows our proposed framework, namely CLUVAE. 
The proposed framework includes four steps:
Firstly, we collect the land-use configurations and the corresponding urban functional zones of a series of target areas;
Secondly, we obtain the embedding of surrounding contexts' features and human guidance respectively.
For the embedding of surrounding contexts' features, we utilize a spatial attributed graph to organize them and employ a graph embedding model to obtain the corresponding embedding.
For the embedding of human guidance, we regard the one-hot vector of the texts of human guidance as the embedding.
Thirdly, we build up a deep robust configuration generation model. 
Specifically, we first concatenate the surrounding contexts'  embedding and human guidance embedding as the condition embedding.
Then, the encoder of the generation model  learns the distribution that reflects the correlation between the condition embedding and the corresponding land-use configuration.
Next, the decoder of the generation model reconstructs land-use configurations and urban functional zones based on the condition embedding and the latent embedding sampled from the previously learned distribution.
Finally, the well-trained decoder is our desired land-use configuration generator. 

\subsection{Finding Target Area and Surrounding Contexts}
How to find target area and surrounding contexts is the cornerstone of our framework.
Specifically, we first collect a list of geographical points that are composed of longitude and latitude coordinates.
Then, we regard each geographical point as the central point to draw the corresponding target area (i.e. one square shape with 1 km$^2$ area).
Later, we find all surrounding contexts of each target area according to the spatial relationships depicted in section \ref{target}.
After that, we collect all target areas and surrounding contexts, and pair them together.
In the following paragraphs, to be convenient, we employ the $k$-th target area and the corresponding surrounding contexts to introduce denotations and all calculation processes.

\subsection{Collecting Land-use Configurations and Discovering Urban Functional Zones}
To build up an AI-based land-use configuration generation model, we need to collect a series of land-use configuration samples.
The collecting process of the $k$-th target area as follows:
we first divide the $k$-th target area into $N \times N$ squares based on latitude and longitude.
Next, we count the number of POIs of each POI category in each square, and organize the numbers of POIs belonging to the same POI category as a matrix according to the squares' position.
After that, we stack different matrices together to form a latitude-longitude-channel tensor.
The format of the tensor as shown in Figure \ref{landuse image} and the $k$-th land-use configuration denoted by $\mathbf{\hat{X}}^{(k)} \in \mathbb{R}^{N \times N \times M}$, where $M$ is the number of POI category.

In addition, urban functional zones provide a rough skeleton for urban planning, which stimulates us to employ them in our generation model for improving generative performance.
Owing to lacking of standard urban functional zones data samples, we utilize the method in \cite{yuan2014discovering} to discover urban functional zones of target areas based on spatio-temporal data.
Limited by space, we only provide the data format of the urban functional zones.
Figure \ref{urban_func} has provided an example of urban functional zones.
For the $k$-th target area, the urban functional zones denoted by $\mathbf{F}^{(k)} \in \mathbb{R}^{N \times N}$, where each value in $\mathbf{F}^{(k)}$ indicates that the corresponding geographical square belongs to which urban functional zone.

\subsection{Extracting Features of the Surrounding Contexts}
The socioeconomic features of surrounding contexts affect the urban planning design of the target area.
For instance, if surrounding contexts have many apartments for people living, the target area should be constructed entertainment POIs (\textit{e.g. Gym, Club}) to improve the vibrancy of all geographical area.
In this paper, we utilize a spatial attributed graph to embrace all features of surrounding contexts, and employ a graph embedding model to convert these features into an embedding.

Firstly, from section \ref{target}, we can find that 
the spatial structural relationship among the surrounding contexts can be regarded as a ring that is a special graph.
For the surrounding contexts of the $k$-th target area, the special graph denoted by $\mathcal{G}^{(k)} = (\mathcal{V}^{(k)},\mathcal{E}^{(k)})$, where $\mathcal{V}^{(k)}$ indicates all vertices of the graph, and one vertex is one surrounding context; 
$\mathcal{E}^{(k)}$ represents all edges of the graph, and each edge indicates the connectivity between any two vertices.

Secondly, we extract explicit features of each vertex from three perspectives: 
(1) \textbf{house price change}, which reflects people's living preference for a geographical area.
Specifically, for each vertex, we first collect the house price in $T$ months.
Then, we obtain the house price change by utilizing the house price of each month to deduct the previous one.
Finally, we collect the house price change of all contexts, denoted by $\mathbf{V} = \{\mathbf{v}_1,\mathbf{v}_2,\dots,\mathbf{v}_8\}$, where $\mathbf{V} \in \mathbb{R}^{8 \times (T-1)}$ and $\mathbf{v}$ is the house price change of a context.
(2) \textbf{POI Ratio}, which describes the distribution of urban functions of one geographical area from the overall perspective.
Specifically, for each vertex, we first count the number of POIs belonging to each POI category.
We then scale the value of each POI category by dividing the total number of POIs in the context, and organize these scaled values as a POI ratio vector.
Finally, we collect the POI ratio vector of all contexts, denoted by $\mathbf{R}=\{\mathbf{r}_1,\mathbf{r}_2,\dots,\mathbf{r}_8\}$, where $\mathbf{R} \in \mathbb{R}^{8 \times M}$
and $\mathbf{r}$ is the POI ratio vector of a context. 
(3)\textbf{Transportation}, which demonstrates the traffic condition of a geographical area. 
Here, we extract features related to transportation from two sides: public transportation and private transportation.
For \textbf{public transportation}, in each context, we collect 5 values: (a) the leaving volume of buses per day; (b) the arriving volume of buses per day; (c) the transition volume of buses per day; (d) the number of bus stops in per square meters; (e) the average price of each bus trip.
The five values are organized as the public transportation feature vector, denoted by $\mathbf{o}$.
Finally, we collect the vectors of all contexts, denoted by $\mathbf{O} = \{ \mathbf{o}_1, \mathbf{o}_2, \dots, \mathbf{o}_8 \}$, where $\mathbf{O} \in \mathbb{R}^{8 \times 5}$.
For \textbf{private transportation}, 
in each context, we also collect 5 values: (a) the leaving volume of taxis per day; (b) the arriving volume of taxis per day; (c) the transition volume of taxis per day; (d) the average velocity of taxis per day; (e) the average commute distance of taxis per day.
The five values are organized as the private transportation feature vector, denoted by $\mathbf{u}$.
Finally, we collect the vectors of all contexts, denoted by $\mathbf{U}=\{ \mathbf{u}_1, \mathbf{u}_2, \dots, \mathbf{u}_8\}$, where $\mathbf{U} \in \mathbb{R}^{8\times 5}$. 

Thirdly, we horizontally concatenate the feature vectors of all contexts together to form the feature set, denoted by  $\mathcal{F}^{(k)}=[\mathbf{V},\mathbf{R},\mathbf{O},\mathbf{U}]$.
$\mathcal{F}^{(k)} \in R^{8\times (M+T+9)}$, where one row indicates all kinds of feature of one vertex. 
Then, we add attributes to the graph $\mathcal{G}^{(k)}$ by matching each row of $\mathcal{F}^{(k)}$ to the corresponding vertex for forming the spatial attributed graph, denoted by $\mathcal{G}^{(k)} = (\mathcal{V}^{(k)},\mathcal{E}^{(k)},\mathcal{F}^{(k)})$.
Finally, a variational graph auto encoder model \cite{kipf2016variational} is used to convert $\mathcal{G}^{(k)}$ into the graph embedding $\mathbf{s}^{(k)}$ that contains not only the spatial structural information among surrounding contexts but also all socioeconomic features.

\begin{figure*}[!t]
    \centering
    \includegraphics[width=0.75\linewidth]{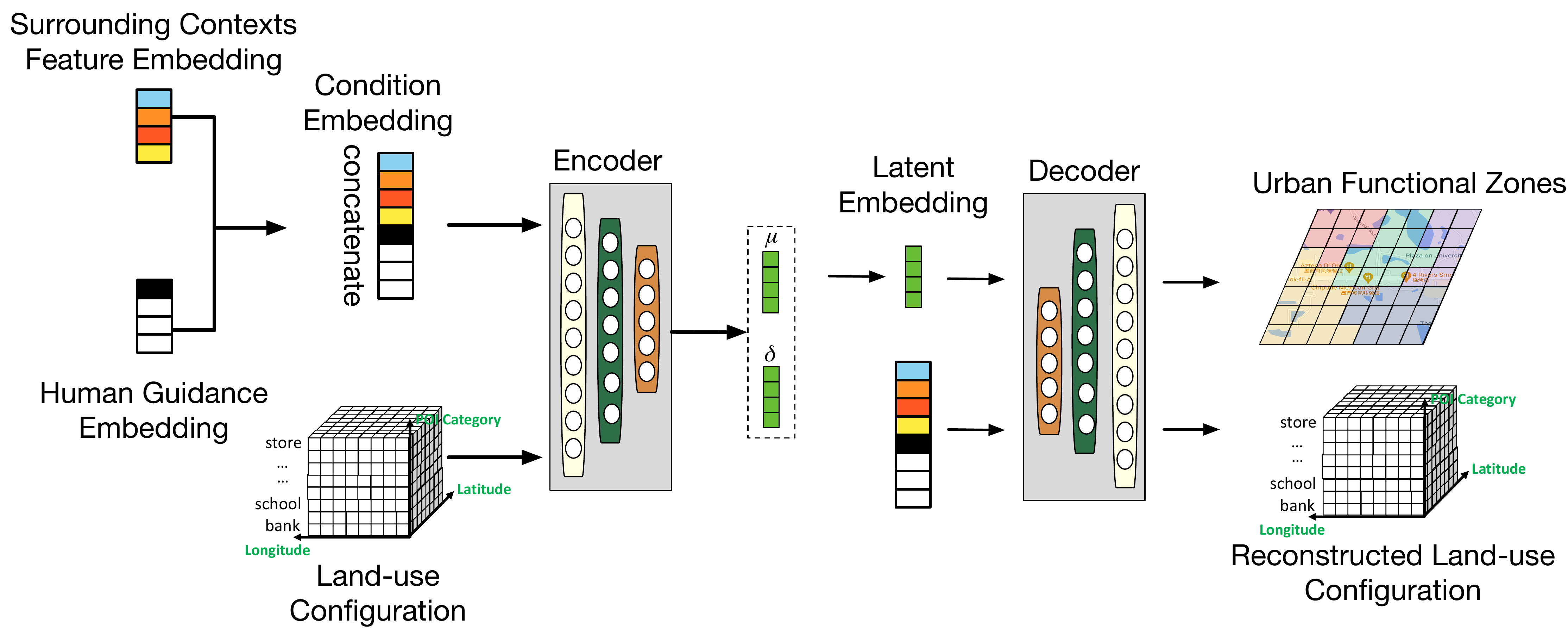}
    \captionsetup{justification=centering}
    \caption{Land-use Configuration Generation Model.}
    \vspace{-0.5cm}
    \label{fig:model_structure}
\end{figure*}

\subsection{Embedding Human Guidance}
Human guidance describes humans' requirements for land-use configuration generation.
Urban planning is a highly personalized task, because different urban planners have different planning ideas, 
meanwhile, different urban planning projects are constrained by different realistic factors. 
So how to make our generated model perceive human guidance is a big challenge.
To overcome this challenge,  we simplify the human guidance into options that can be selected by urban planners according to their needs.
For instance, if urban planners have greenery rate requirements for producing a land-use configuration, they will have different greenery levels to select.
After the selection, our model can produce the corresponding result based on planners' choice.
Here, we use the one hot vector of the selected level label as the embedding of human guidance.
The human guidance for the $k$-th target area, denoted by $\mathbf{i}^{(k)}$.

Ultimately, for the $k$-th target area, we have obtained the land-use configuration $\mathbf{\hat{X}}^{(k)}$, the urban functional zones $\mathbf{F}^{(k)}$, the embedding of surrounding contexts' features $\mathbf{s}^{(k)}$, and the embedding of the human guidance $\mathbf{i}^{(k)}$.
Then, we organize them as a group $[\mathbf{\hat{X}}^{(k)},\mathbf{F}^{(k)},\mathbf{s}^{(k)}, \mathbf{i}^{(k)}]$, which is one data sample learned by our following model structure.

\subsection{Conditional Land-use Variational Autoencoder}
\label{cluvae}
Our model aims to find the mapping function between land-use configurations and the corresponding conditions (\textit{i.e. surrounding contexts and human instructions}).
To achieve this goal, our framework is implemented through an encoder-decoder paradigm.

The encoder part firstly converts the land-use configuration $\mathbf{\hat{X}}^{(k)}$ and the conditions ($\mathbf{s}^{(k)}, \mathbf{i}^{(k)}$) into a latent embedding vector $\mathbf{z}^{(k)}$.
During the conversion process, 
we utilize a variational Gaussian embedding component to find the conditional distribution $p(\mathbf{z}^{(k)} | \mathbf{\hat{X}}^{(k)}, \mathbf{s}^{(k)}, \mathbf{i}^{(k)})$.
Owing to we assume the conditional distribution belongs to the family of normal distribution, the encoder part only needs to estimate the parameters (\textit{i.e. mean $\bm{\mu}^{(k)}$, variance $\bm{\delta}^{(k)}$}) of the distribution and produce the latent embedding $\mathbf{z}^{(k)}$.
Specifically,  we first concatenate the embedding of surrounding contexts' features $\mathbf{s}^{(k)}$ and the embedding of human instructions $\mathbf{i}^{(k)}$ as the condition embedding $\mathbf{c}^{(k)}$.
Then we flatten the land-use configuration $\mathbf{\hat{X}}^{(k)}$ into a one-dimensional vector, denoted by $\mathbf{x}^{(k)}$.
Later, we input $\mathbf{c}^{(k)}$ and $\mathbf{x}^{(k)}$ into two fully connected layers respectively to learn the parameters of the conditional distribution $p(\mathbf{z}^{(k)} | \mathbf{x}^{(k)} , \mathbf{c}^{(k)}) $, where $p(\mathbf{z}^{(k)} | \mathbf{x}^{(k)} , \mathbf{c}^{(k)}) $ equals $p(\mathbf{z}^{(k)} | \mathbf{\hat{X}}^{(k)}, \mathbf{s}^{(k)}, \mathbf{i}^{(k)})$.
After that, we utilize a reparametrization technique to sample latent embedding $\mathbf{z}^{(k)}$ from the learned distribution.
The whole calculation process can be formulated as follows:
\begin{equation}
\left\{
             \begin{array}{lr}
             \mathbf{c}^{(k)} = \text{Concat}(\mathbf{s}^{(k)},\mathbf{i}^{(k)}),\\
             \mathbf{x}^{(k)} = \text{Flatten}(\mathbf{\hat{X}}^{(k)}),\\
             \bm{\mu}^{(k)} = \text{Fully}_1(\mathbf{x}^{(k)},\mathbf{c}^{(k)}),
             \\
             \bm{\delta}^{(k)} = \text{Fully}_2(\mathbf{x}^{(k)},\mathbf{c}^{(k)}),
             \\
             \mathbf{z}^{(k)}=\bm{\mu}^{(k)}+\bm{\delta}^{(k)} \times \epsilon.
              &  
             \end{array}
\right.
\label{equ:encoder}
\end{equation}
The last line of equation \ref{equ:encoder} is the reparameterization trick, where $\epsilon$ sampled from standard normal distribution $N(0,1)$.
Next, we input $\mathbf{z}^{(k)}$ and $\mathbf{c}^{(k)}$ into the decoder part to conduct the decoding step.

The decoder part reconstructs $\mathbf{\hat{X}}^{(k)}$ and constructs the corresponding urban functional zones $\mathbf{F}^{(k)}$ based on $\mathbf{z}^{(k)}$ and $\mathbf{c}^{(k)}$. 
Inspired by \cite{ding2020guided}, the multi-head decoder in VAE can improve model performance by adding certain constraints.
Here, we regard urban functional zones as the generated  constraints in the decoder for producing  reasonable land-use configurations.
Specifically, we first input $\mathbf{z}^{(k)}$ and $\mathbf{c}^{(k)}$ a fully connected layer activated by Relu function to get the reconstructed land-use configuration $\mathbf{\hat{\ddot{X}}^{(k)}}$.
Next, we input  $\mathbf{z}^{(k)}$ and $\mathbf{c}^{(k)}$ into another fully connected layer activated by sigmoid function $\sigma$ to construct urban functional zones $\mathbf{\ddot{F}}^{(k)}$.
The calculation process can be formulated as:
\begin{equation}
\left\{
             \begin{array}{lr}
             \mathbf{\hat{\ddot{X}}^{(k)}} = \text{Relu}(\text{Fully}_3(\mathbf{z}^{(k)},\mathbf{c}^{(k)})),\\
             \mathbf{\ddot{F}^{(k)}} =\sigma(\text{Fully}_4(\mathbf{z}^{(k)},\mathbf{c}^{(k)})).
              &  
             \end{array}
\right.
\label{equ:decoder}
\end{equation}

There are three learning objectives during the training phase:
(1) minimizing the difference between  $\mathbf{\hat{\ddot{X}}^{(k)}}$ and $\mathbf{\hat{X}^{(k)}}$, denoted by $\mathcal{L}_X$;
(2) owing to we assume the prior distribution of $p(\mathbf{z}^{(k)} | \mathbf{x}^{(k)} , \mathbf{c}^{(k)}) $ is $q(\mathbf{z}^{(k)})$ that is a standard normal distribution $N(0,1)$, we need to minimize the gap between the two distributions, denoted by $\mathcal{L}_p$;
(3) minimizing the difference between  $\mathbf{\ddot{F}^{(k)}}$ and $\mathbf{F^{(k)}}$, denoted by $\mathcal{L}_F$;
The final loss $\mathcal{L}$ is the combination of $\mathcal{L}_X$, $\mathcal{L}_p$, and $\mathcal{L}_F$, which can be formulated as follows:
\begin{equation}
\left\{
             \begin{array}{lr}
             \mathcal{L}_X= \frac{\sum^{K}_{k=1}( \mathbf{\hat{X}^{(k)}} - \mathbf{\hat{\ddot{X}}^{(k)}})^2}{K}
             \\
             \mathcal{L}_p= \frac{\sum^{K}_{k=1}\text{KL}[p(\mathbf{z}^{(k)}|\mathbf{x}^{(k)} , \mathbf{c}^{(k)}) || q(\mathbf{z}^{(k)})]}{K} 
             \\
             \mathcal{L}_F= - \frac{\sum_{k=1}^{K} \mathbf{F}^{(k)}\times \text{log}(\mathbf{\ddot{F}}^{(k)})}{K}
             \\
             \mathcal{L} = \mathcal{L}_X + \mathcal{L}_p + \lambda \times \mathcal{L}_F
              &  
             \end{array}
\right.
\label{equ:decoder}
\end{equation}
where KL represents the Kullback-Leibler divergence between $p(.)$ and $q(.)$;
$\lambda$ controls the strength of the generated constraints (\textit{i.e. urban functional zones}) during the model learning process.
The value range of $\lambda$ is $[0 \sim 1]$.

After the generation model converges, the decoder is the land-use configuration generator.
During the testing phase, we first sample a latent embedding from the learned distribution based on the condition embedding.
Then, the decoder takes the sampled latent embedding and the condition embedding as input, and outputs the generated land-use configuration.

\subsection{Difference from Prior Literature}
With the development of artificial intelligence (AI), how to utilize AI models to improve and accelerate urban planning attracts many researchers.
However, there are certain limitations in prior literature: 1) the automated urban planning process cannot be personalized based on human's requirements; 2) the generated urban planning solutions are unreasonable;
3) the available standard urban planning solution is sparse.
Our study aims to overcome the three limitations.
In our framework, the human guidance for urban planning is regarded as a condition input of  the land-use configuration generation model.
In addition, We add generated constraints (\textit{i.e., urban functional zones}) in the generation model for improving the reasonability of generated results.
Moreover, we employ a variational Gaussian embedding mechanism to make the model learning process better.
This is how this study differentiates from advanced prior literature.

\section{Experimental Results}
We presented extensive experiments to answer:
\textbf{Q1.} Does our method (CLUVAE) outperform other baseline models?
\textbf{Q2.} Does the variational Gaussian embedding mechanism make the model learning process stable?
\textbf{Q3.} Is each technical component of CLUVAE necessary to generate better land-use configurations?
\textbf{Q4.} 
We split the target area into $N\times N$ squares to collect land-use configurations. 
How does the square size influence  configuration generation?
\textbf{Q5.} What do the generated land-use configurations look like? What are the differences between generated configurations and real configurations?

\subsection{Experimental Setup}

\subsubsection{Data Description}
The datasets used in the experiments include: 
\textbf{(1) Residential Communities:}
 are crawled from soufun.com and contains 2990 residential communities in Beijing. 
 These communities are regarded as target areas to collect data samples for model learning and each community has its green rate record.
\textbf{(2) POIs:} include 328,668 POIs in Beijing. Each POI is associated with latitude, longitude, POI categories.  Table \ref{poi_lists} shows the 20 POI categories. 
\textbf{(3) Taxi Trajectories:}  are collected from a taxi company in Beijing. Each trajectory contains a trip ID, a travel distance, travel times, average speed, pick-up and drop-off times, pick-up and drop-off points. 
\textbf{(4) Public Transportation:} includes bus transactions in Beijing from 2012 to 2013 and contains 718 bus lines, 1734247 bus trips.
\textbf{(5) Housing Prices:} are collected from soufun.com.
\textbf{(6) Mobile Checkins: }which are collected from weibo.com. Each check-in record includes longitude, latitude, check-in time and place.

\begin{table}[htbp]
\vspace{-0.2cm}
\small
\centering
\setlength{\abovecaptionskip}{0.cm}
\caption{POI categories}
\setlength{\tabcolsep}{1mm}{
\begin{tabular}{cccccc}  
\toprule
 code  & POI category & code & POI category  \\  
\midrule       
  0  & road & 10 & tourist attraction \\
  1 & car service & 11 & real estate \\
  2 & car repair & 12 & government place \\
  3 & motorbike service & 13 & education \\
  4 & food service & 14 & transportation \\
  5 & shopping & 15 & finance\\
  6 & daily life service & 16 & company\\
  7 & recreation service & 17 & road furniture\\
  8 & medical service & 18 & specific address \\
  9 & lodging  & 19 & public service\\
\bottomrule
\end{tabular}}
\label{poi_lists}
\vspace{-0.2cm}
\end{table}

\subsubsection{Evaluation Metrics}
We categorized the collected land-use configurations into five levels according to their greenery rate.
Green level labels are regarded as human guidance in our experiments.
The land-use configurations of different green levels come from different distributions.
To evaluate the model performance, we measured the distance between the distribution of generated configurations and the distribution of original configurations under different green level settings.
Let $w_j$ denote the weight of the green level $j$, which is the number of land-use configurations belonging to the green level $j$; 
$Y_j$ denote the distribution of original land-use configurations of green level $j$;
$\hat{Y}_j$ denote the distribution of generated land-use configurations of green level $j$;
there are four evaluation metrics in our work: 
1) \textbf{Average Kullback-Leibler (KL) Divergence\footnote{https://en.wikipedia.org/wiki/Kullback-Leibler\_divergence}:}  
$\text{AVG\_KL} = \frac{\sum_{j=1}^{5}w_j\cdot KL(Y_j,\hat{Y}_j) }{\sum_{j=1}^{5}w_j}. $ 
2) \textbf{Average Jensen-Shannon (JS) Divergence\footnote{https://en.wikipedia.org/wiki/Jensen-Shannon\_divergence}:} 
$\text{AVG\_JS} =  \frac{\sum_{j=1}^{5}w_j\cdot JS(Y_j,\hat{Y}_j) }{\sum_{j=1}^{5}w_j}.$
3) \textbf{Average Hellinger Distance (HD) \footnote{https://en.wikipedia.org/wiki/Hellinger\_distance}:} $\text{AVG\_HD} =  \frac{\sum_{j=1}^{5}w_j\cdot HD(Y_j,\hat{Y}_j) }{\sum_{j=1}^{5}w_j}.$
4)  \textbf{ Average Cosine Distance (Cos)\footnote{https://en.wikipedia.org/wiki/Cosine\_similarity}:} 
$ \text{AVG\_Cos} =  \frac{\sum_{j=1}^{5}w_j\cdot Cos(Y_j,\hat{Y}_j) }{\sum_{j=1}^{5}w_j}.$
The lower the metric value is, the better performance the model exhibits.

\begin{figure*}[hbtp!]
	\subfigure[AVG\_KL]{\includegraphics[width=4.35cm]{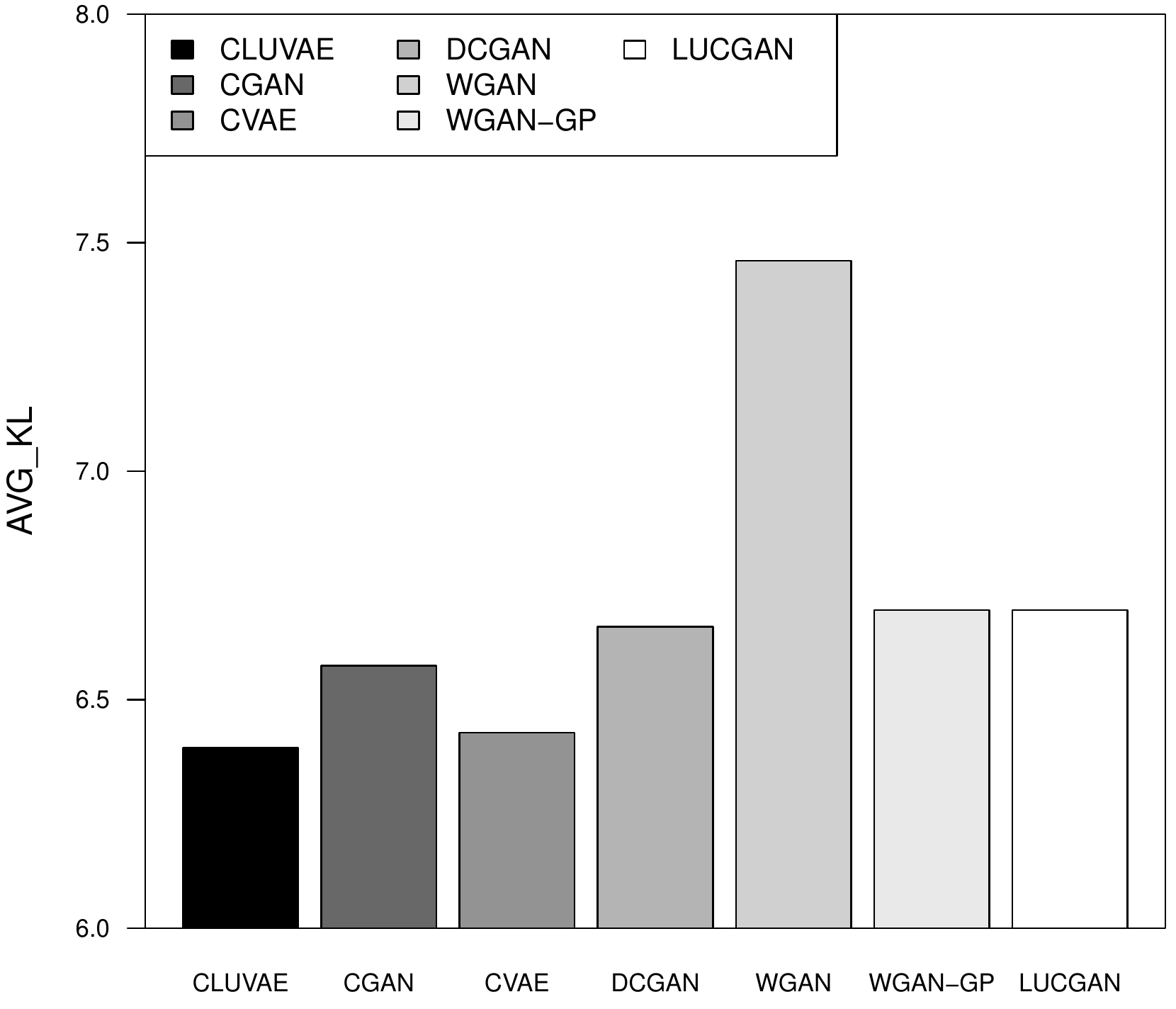} \label{fig:avg_kl}}
	\subfigure[AVG\_JS]{\includegraphics[width=4.35cm]{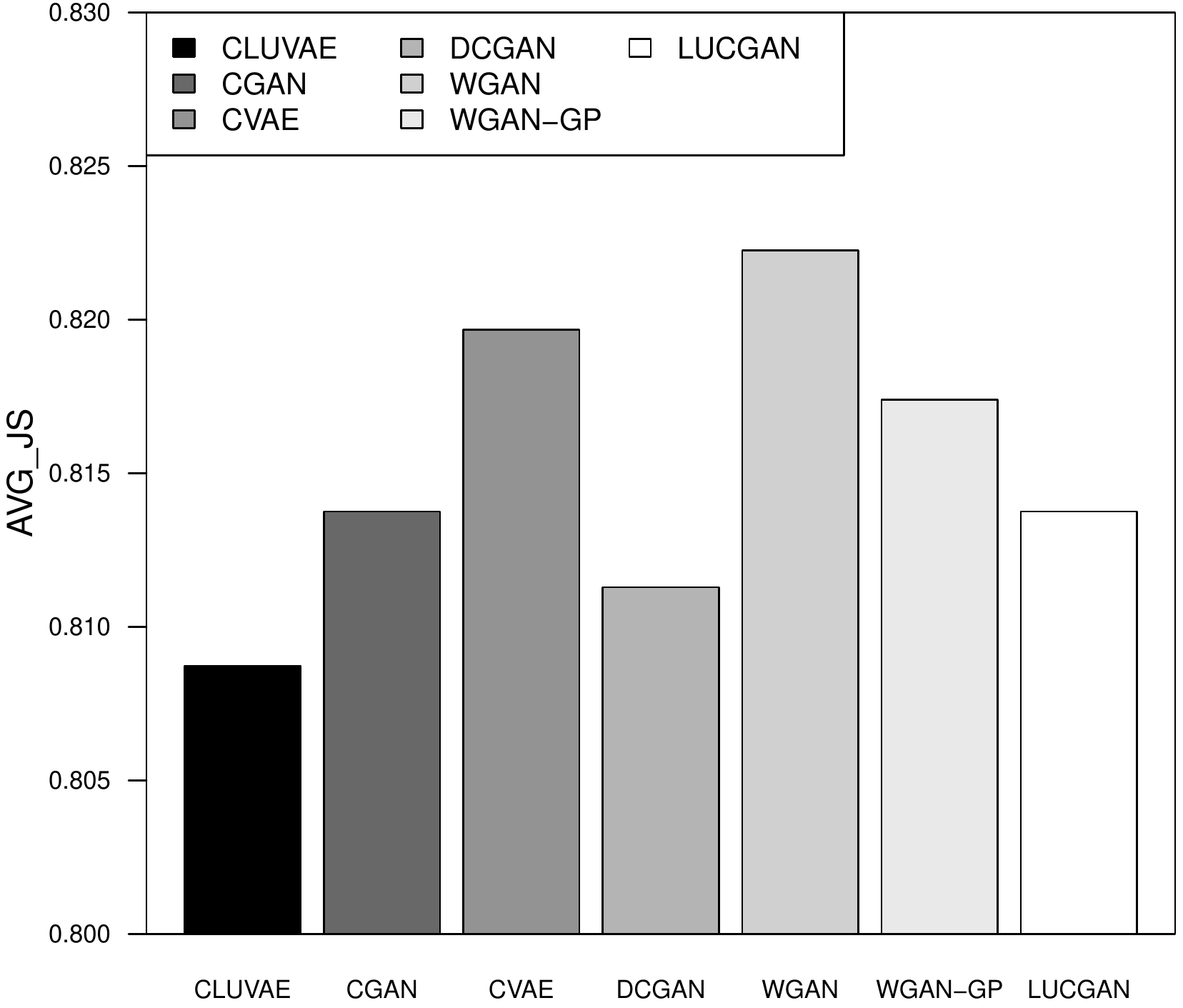}\label{fig:avg_js}}
	\subfigure[AVG\_HD]{\includegraphics[width=4.35cm]{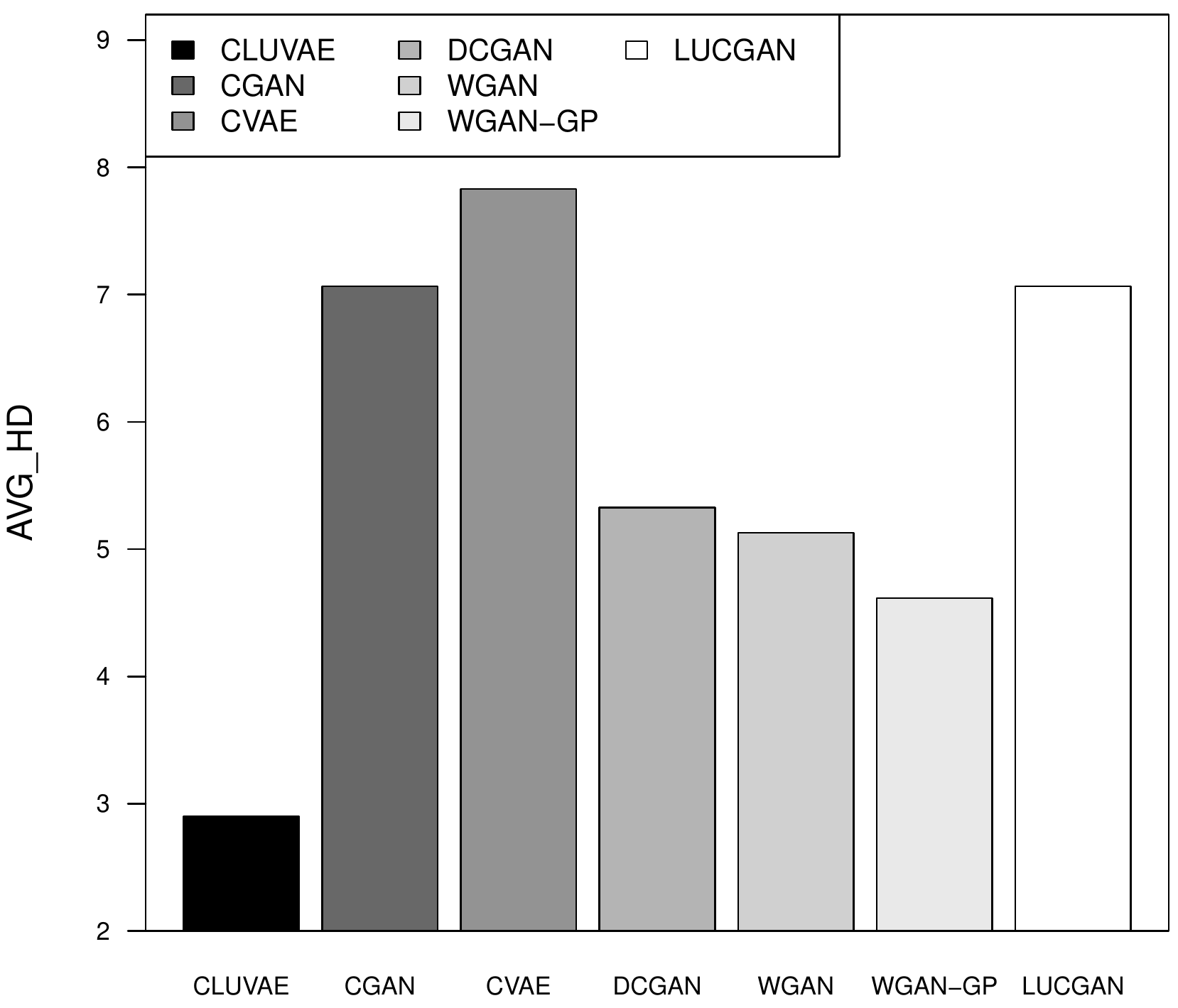}\label{fig:avg_hd}}
	\subfigure[AVG\_Cos]{\includegraphics[width=4.35cm]{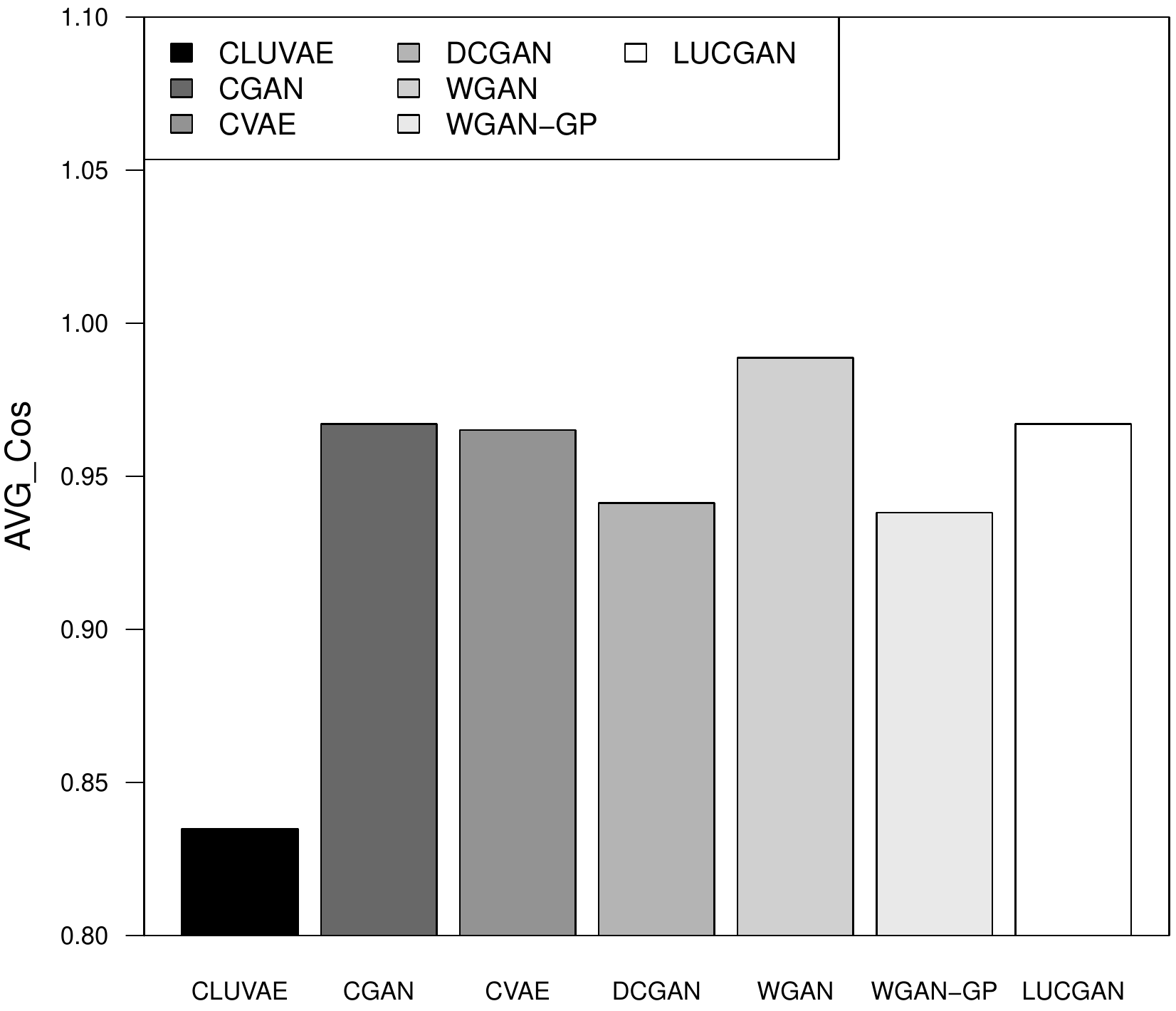}\label{fig:avg_Cos}}	
	\vspace{-0.1cm}
	\caption{Overall Performance in terms of all evaluation metrics.}
	\label{fig:overall performance}
	\vspace{-0.3cm}
\end{figure*}

\begin{figure*}[hbtp!]
	\subfigure[AVG\_KL]{\includegraphics[width=4.35cm]{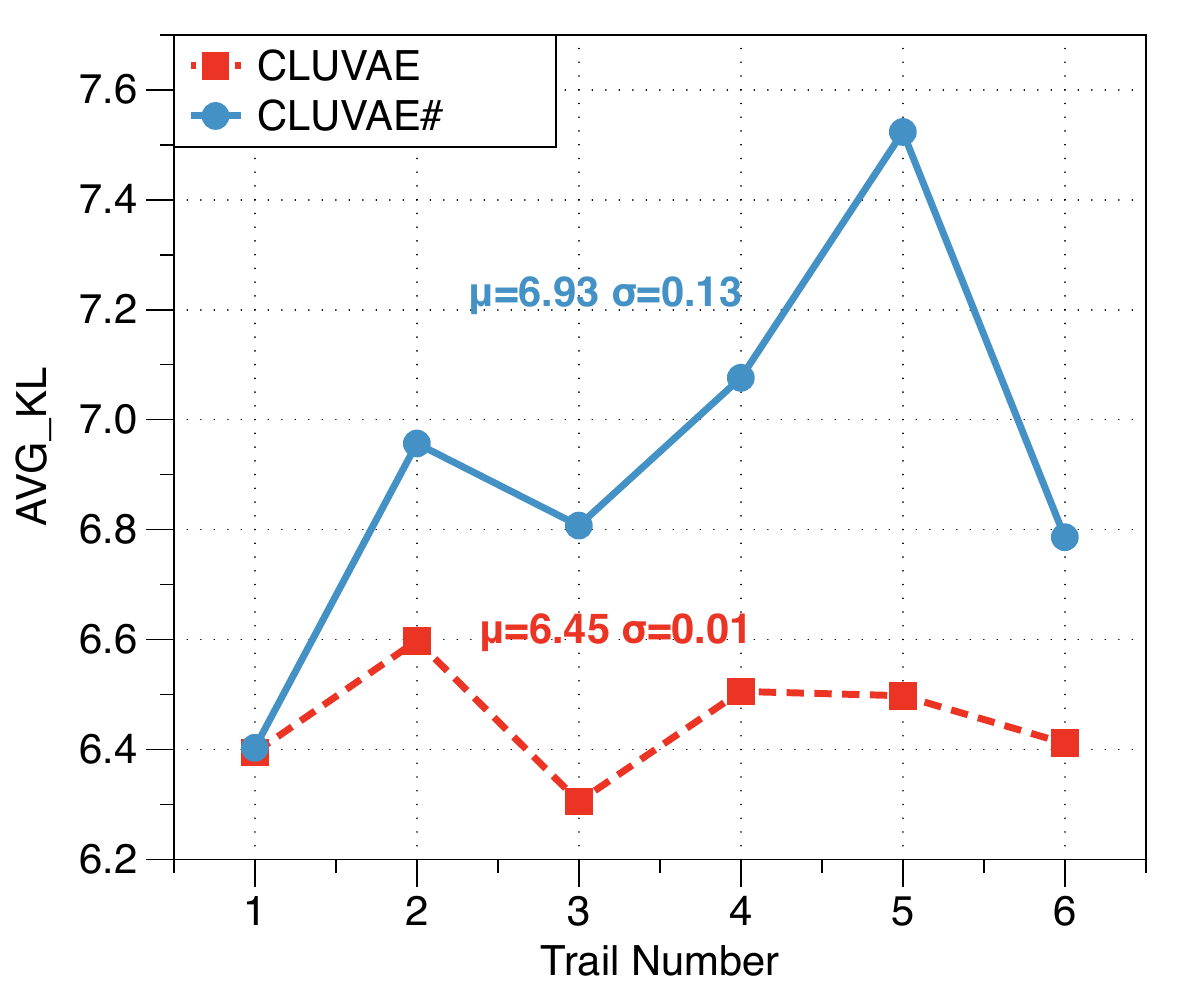} \label{fig:rb_avg_kl}}
	\subfigure[AVG\_JS]{\includegraphics[width=4.35cm]{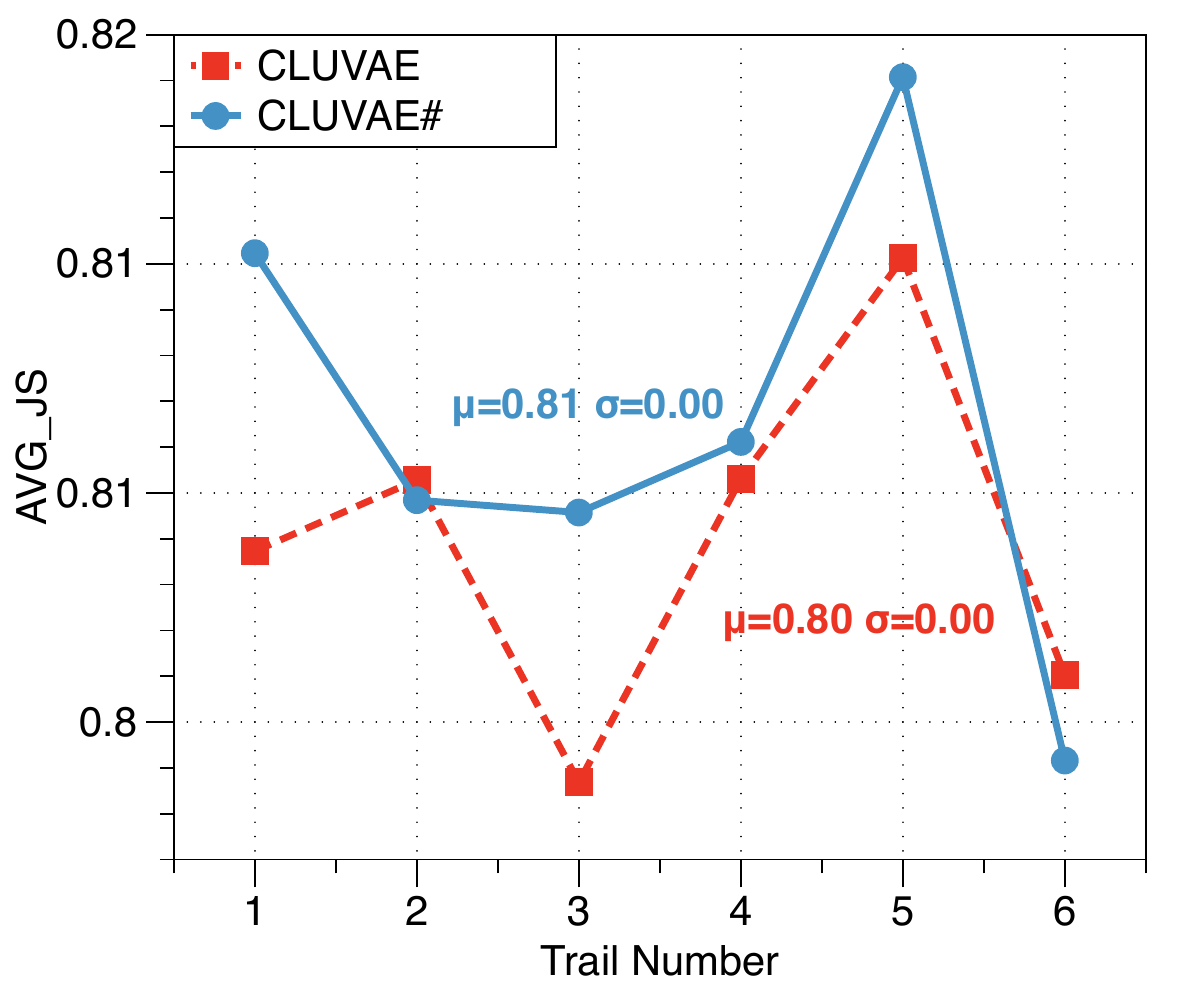}\label{fig:rb_avg_js}}
	\subfigure[AVG\_HD]{\includegraphics[width=4.35cm]{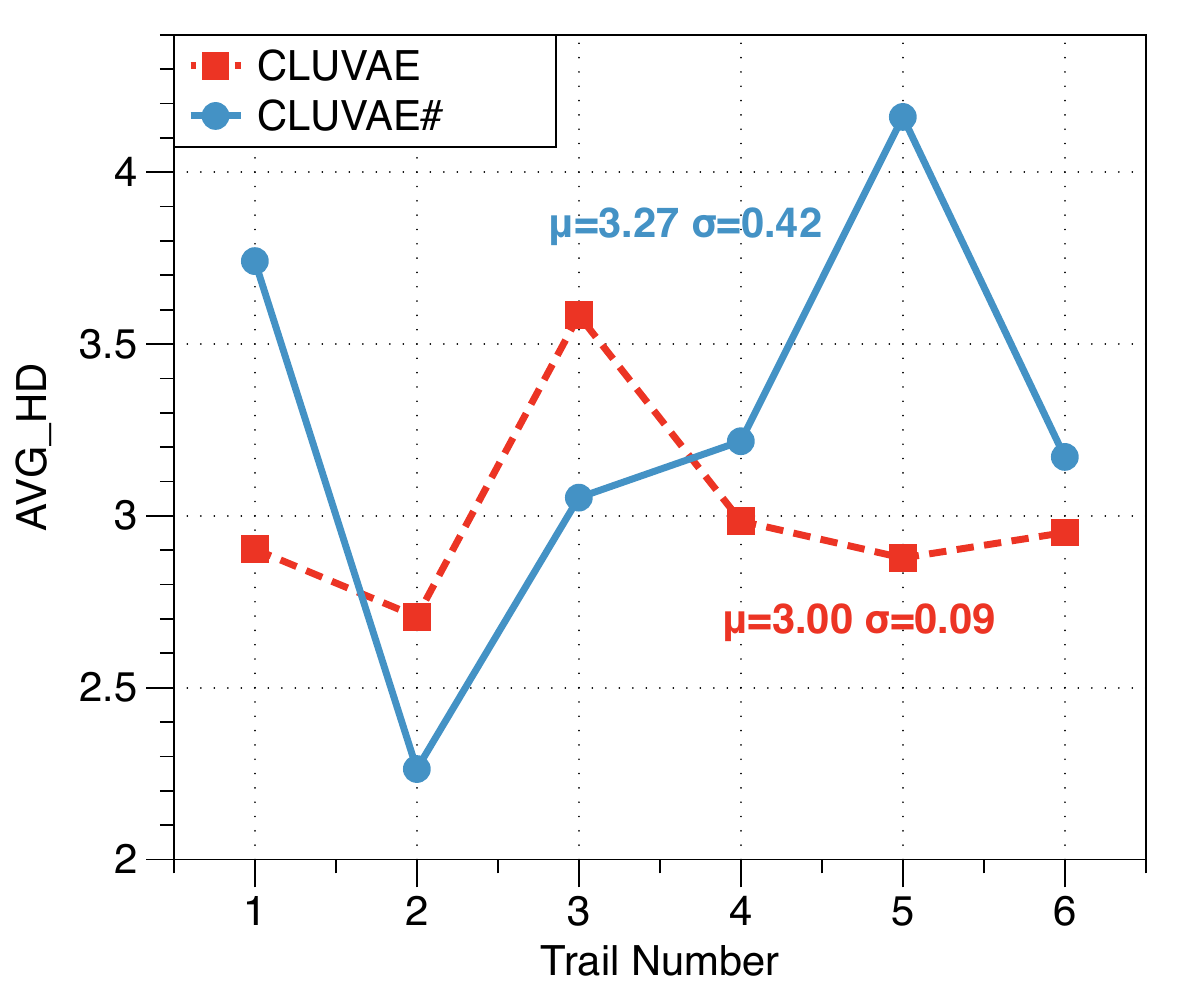}\label{fig:rb_avg_hd}}
	\subfigure[AVG\_Cos]{\includegraphics[width=4.35cm]{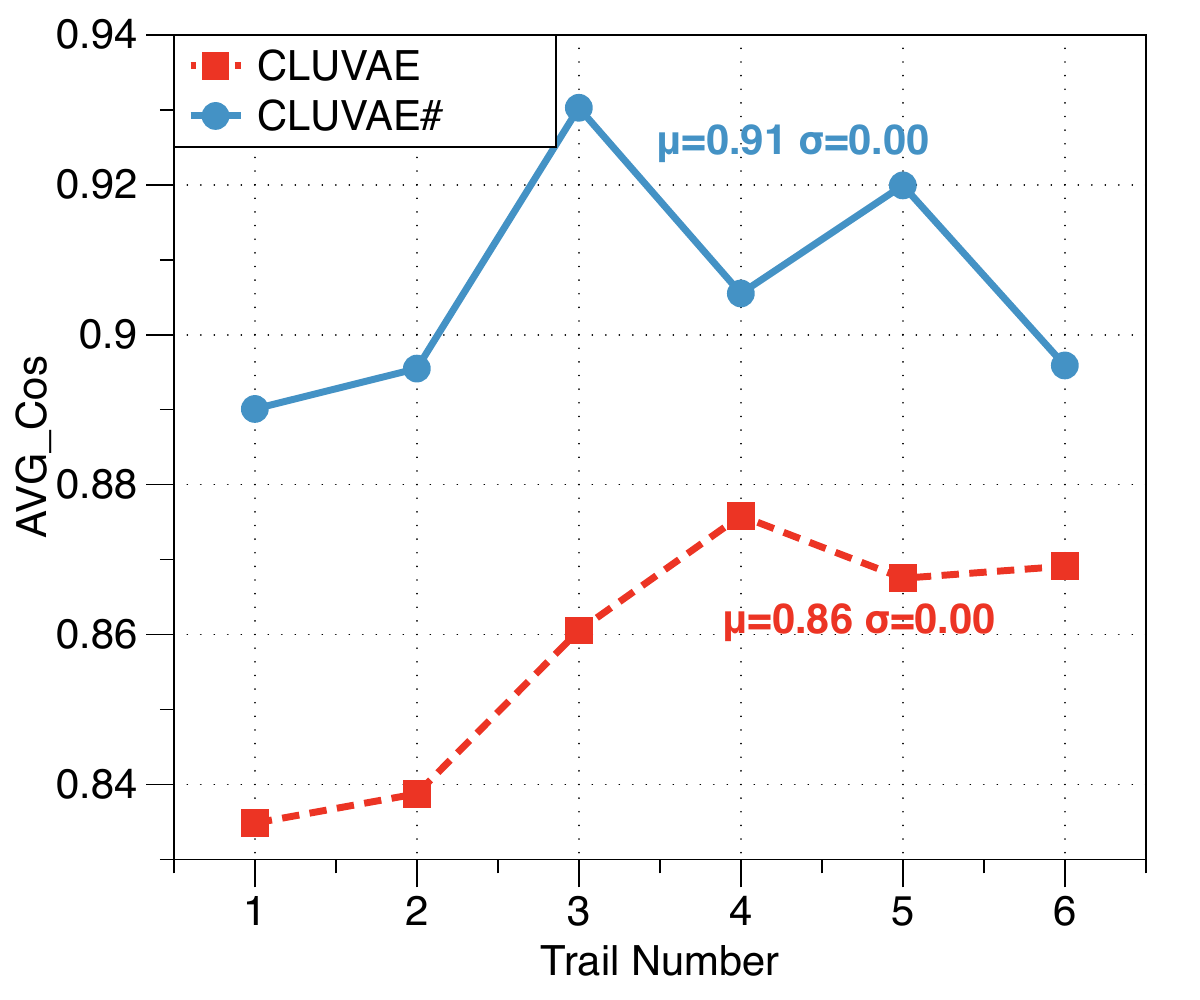}\label{fig:rb_avg_Cos}}	
	\vspace{-0.1cm}
	\caption{Effectiveness check for variational module of CLUVAE in terms of all evaluation metrics.}
	\label{fig:effect_check}
	\vspace{-0.6cm}
\end{figure*}

\subsubsection{Baseline Methods}

We compared our proposed framework (CLUVAE) against the following baseline models:
\textbf{CGAN:}~\cite{mirza2014conditional} is the conditional version of generative adversarial nets (GAN). 
Compared with classical GAN, CGAN adds a conditional input to the generator and discriminator respectively.
\textbf{CVAE:}~\cite{sohn2015learning}  is another conditional generative model, which is an extension of variational auto encoder.
\textbf{DCGAN:}~\cite{radford2015unsupervised} is a classical image generative model, which utilizes convolutional and convolutional transpose layers in the generator and discriminator respectively.    
\textbf{WGAN:}~\cite{pmlr-v70-arjovsky17a} improves the stability of learning of traditional GAN, and makes the learning curve of GAN become meaningful.
\textbf{WGAN-GP:}~\cite{gulrajani2017improved} is an enhanced WGAN, which utilizes gradient penalty to replace clipping weights of WGAN.
\textbf{LUCGAN:}~\cite{wang2020reimagining} is designed for land-use configurations specifically, which can generate configurations automatically based on surrounding contexts' socioeconomic activities.
Besides, to validate the necessity of each component of CLUVAE, we developed four internal variants of CLUVAE: 
\textbf{CLUVAE*}, which removes the embedding of human guidance in condition embedding;
\textbf{CLUVAE$'$}, which removes the embedding of surrounding contexts' features in condition embedding.
\textbf{CLUVAE-}, which removes the generated constraints of CLUVAE;
\textbf{CLUVAE\#},  which removes the variational component and only uses a single auto-encoder as the basic framework of CLUVAE. 


\subsubsection{Environmental Setting} We conducted all experiments in the Ubuntu 18.04.3 LTS operating system, plus Intel(R) Core(TM) i9-9920X CPU@ 3.50GHz, 1 way SLI Titan RTX and 128GB of RAM, with the framework of Python 3.7.4, Tensorflow 2.0.0.

\subsubsection{Hyperparameters and Reproducibility}
In our experiments, for the land-use configuration model as described in Section \ref{cluvae}, the encoder part has two fully connected layers; the decoder part has two branches and each branch is composed of two fully connected layers.
We randomly shuffle our processed dataset and split it into two independent sets according to the index of the target area: the prior 90$\%$ is the training set and the remaining 10$\%$ is the testing set.
During the model training process, the training epochs is set to 50, the Adaptive Moment Estimation (Adam) optimizer is used to optimize the model with learning rate 0.0001, and the value of $\lambda$ used in total loss $\mathcal{L}$ is set to 0.55.
To get the objective experimental results, we set the same value for these basic hyperparameters in other baseline models.

\subsection{Overall Performance \textbf{Q1}}
To validate the effectiveness of CLUVAE, we compared the generated performance of CLUVAE against the baseline methods in terms of AVG\_KL, AVG\_JS, AVG\_HD, and AVG\_Cos.
Figure \ref{fig:overall performance} shows the CLUVAE outperforms other baseline methods in terms of all evaluation metrics. 
This observation indicates that CLUVAE is capable of generating land-use configurations automatically based on the features of surrounding contexts and human instructions.
Meanwhile, compared with other models, CLUVAE captures more inner characteristics of land-use configurations with different green levels.
In summary, this experiments proves the effectiveness of CLUVAE compared with other baseline models. 

\subsection{Effectiveness check of  variational component \textbf{Q2}}
To resolve data sparsity and improve the robustness of our framework, we added a variational module into CLUVAE.
Thus, we designed the experiment to validate that whether or not the variational module achieves our expectations.
We ran CLUVAE and CLUVAE\# six times respectively with random initialization and recorded the change of each evaluation metric.
Figure \ref{fig:effect_check} shows the changes of all metrics under different models.
Meanwhile, to observe the performance and stability more clearly, we provided the mean and variance of the change of each metric in each subfigure.
Figure \ref{fig:effect_check} shows CLUVAE outperforms CLUVAE\# in terms of all metrics.
A possible explanation for the observation is that the variational module helps CLUVAE capture the characteristics of the distribution of land-use configurations further.
In addition, compared with mean and variance, another observation is that CLUVAE is more stable and robust than CLUVAE\#.
A potential interpretation for the observation is that the function of the variational module is equivalent to data augmentation, which enlarges data samples. 
In summary, this experiment proves the variational module of CLUVAE is vital and effective.

\begin{figure*}[hbtp!]
	\vspace{-0.3cm}
	\subfigure[AVG\_KL]{\includegraphics[width=4.35cm]{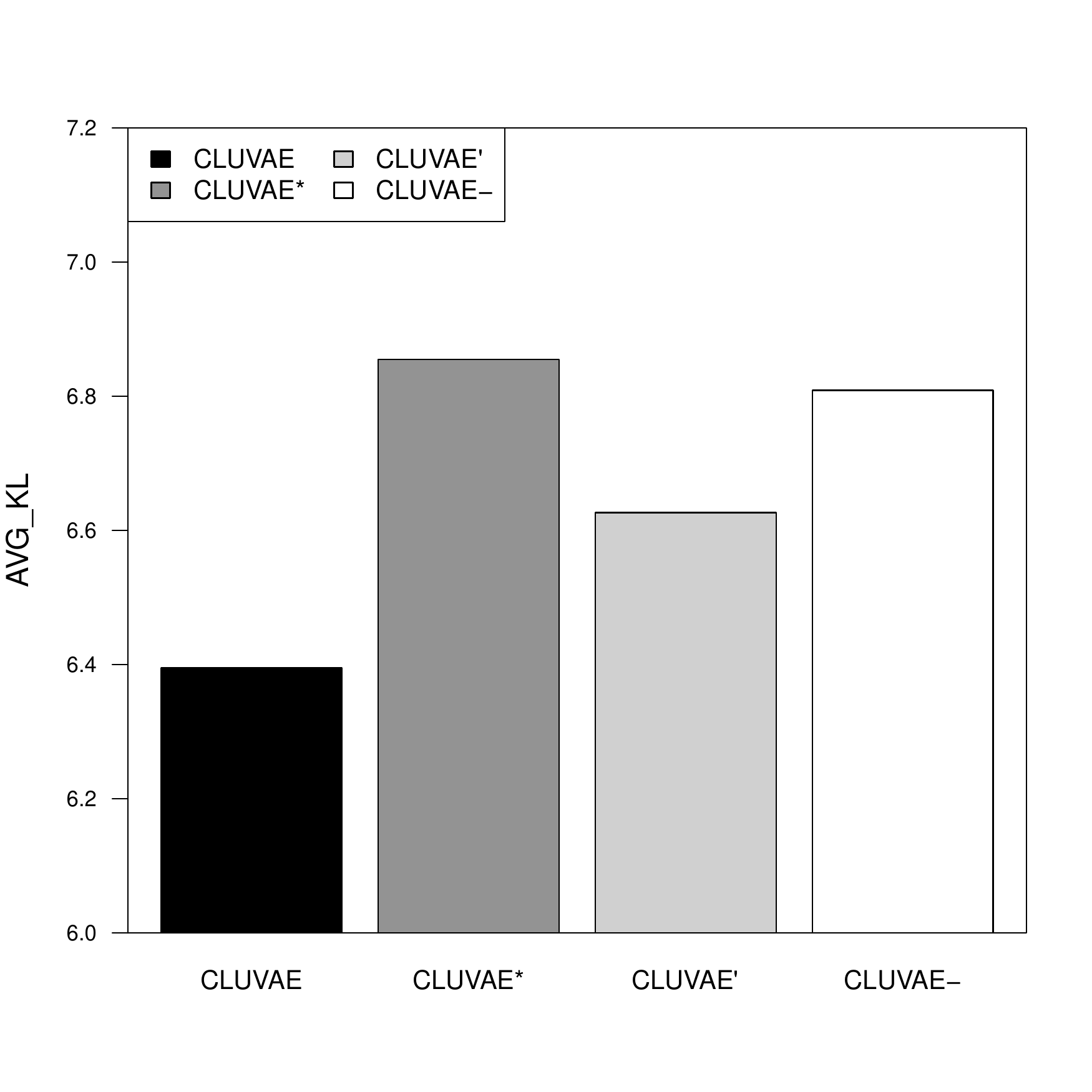} \label{fig:ab_avg_kl}
	}
	\subfigure[AVG\_JS]{\includegraphics[width=4.35cm]{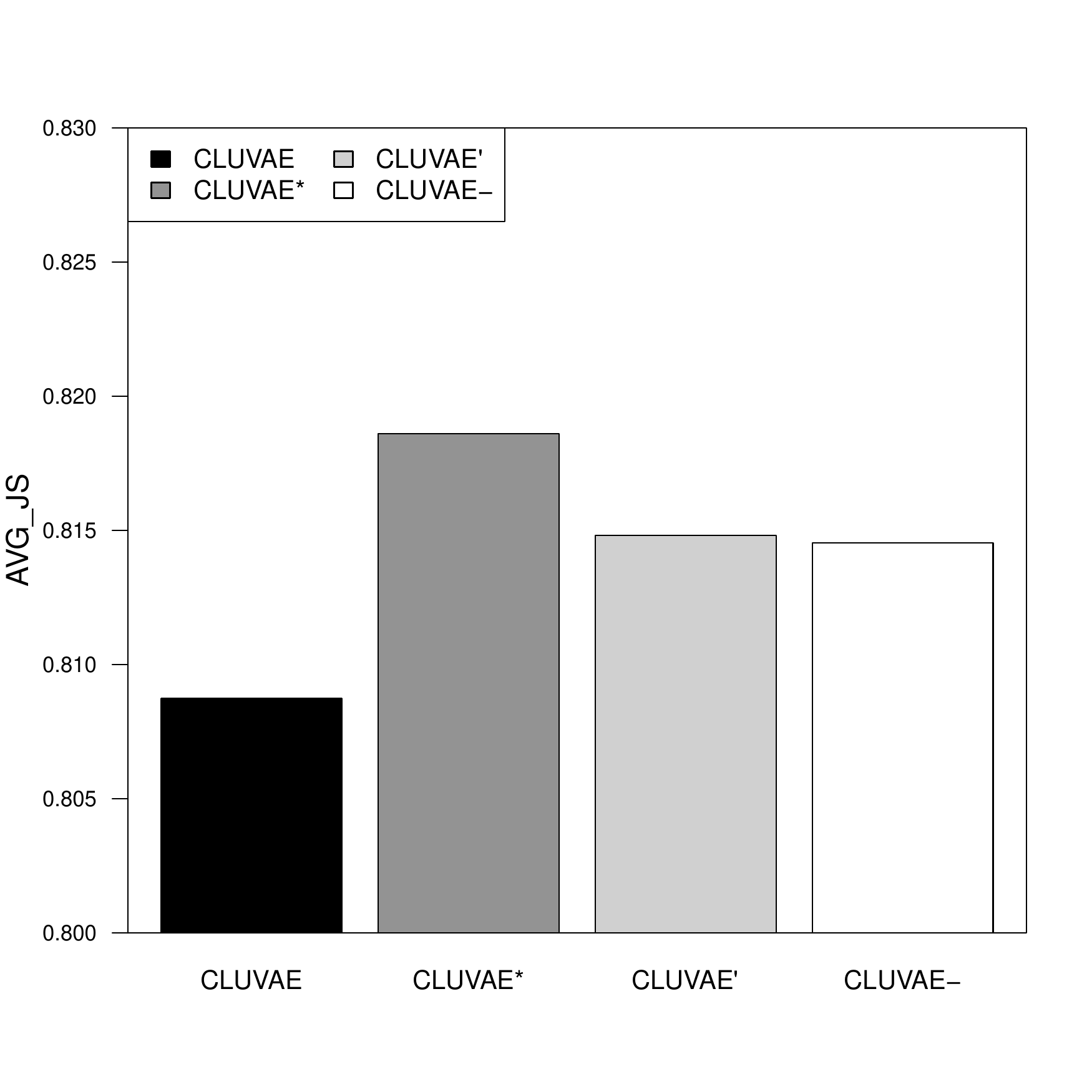}\label{fig:ab_avg_js}}
	\subfigure[AVG\_HD]{\includegraphics[width=4.35cm]{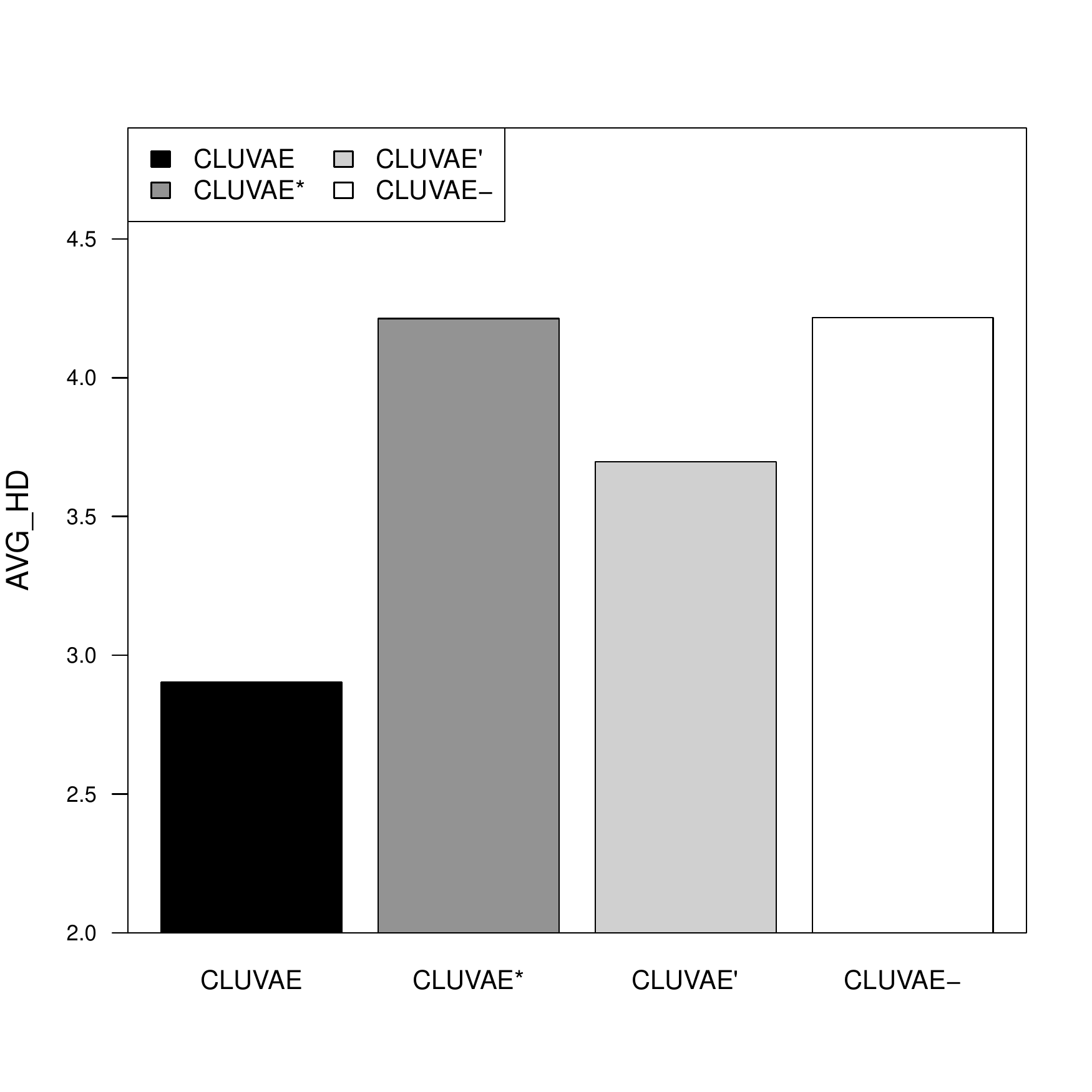}\label{fig:ab_avg_hd}}
	\subfigure[AVG\_Cos]{\includegraphics[width=4.35cm]{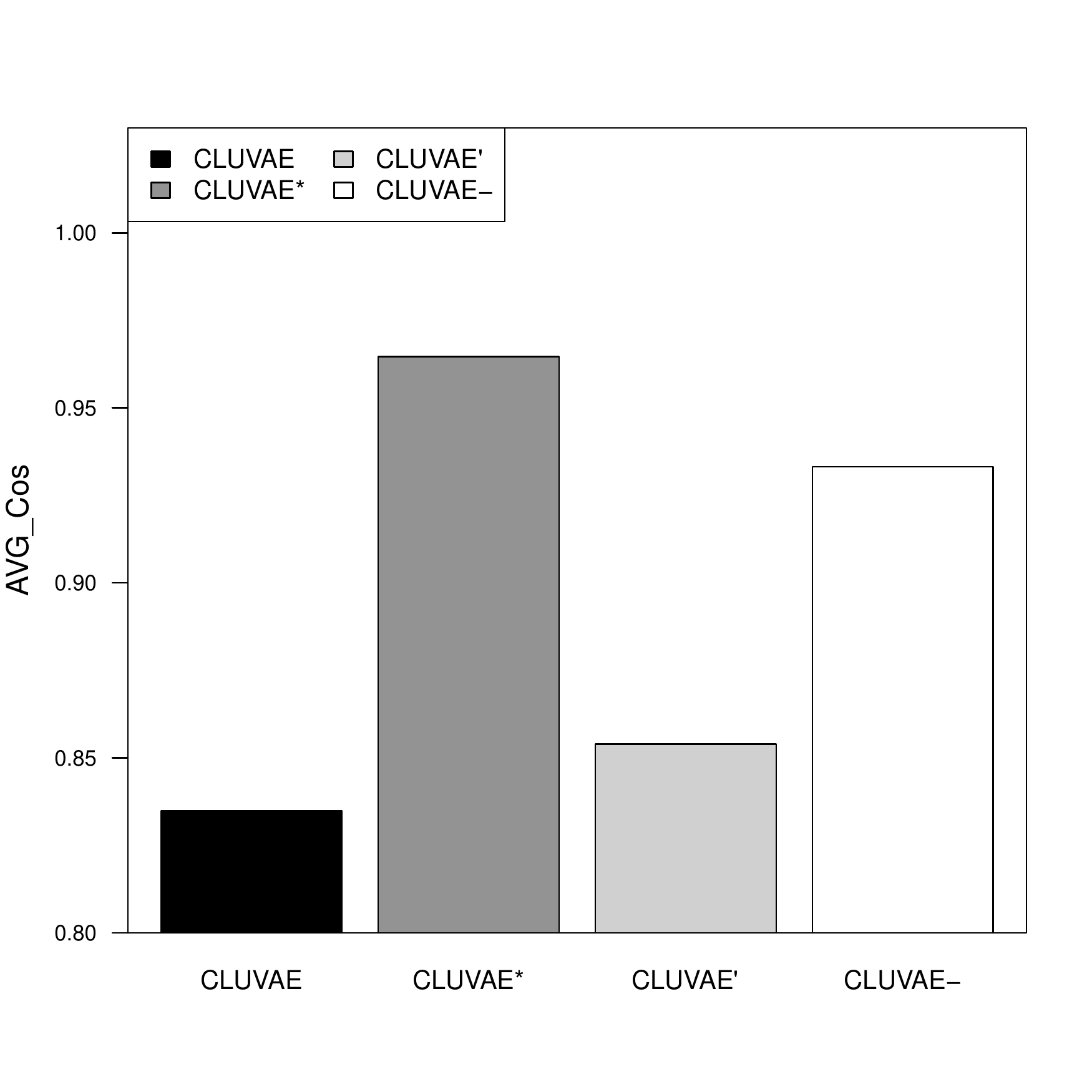}\label{fig:ab_avg_Cos}}	
	\vspace{-0.25cm}
	\caption{Ablation study for CLUVAE in terms of all evaluation metrics.}
	\label{fig:ablation_study}
	\vspace{-0.45cm}
\end{figure*}

\begin{figure*}[hbtp!]
	\vspace{-0.3cm}
	\subfigure[AVG\_KL]{\includegraphics[width=4.35cm]{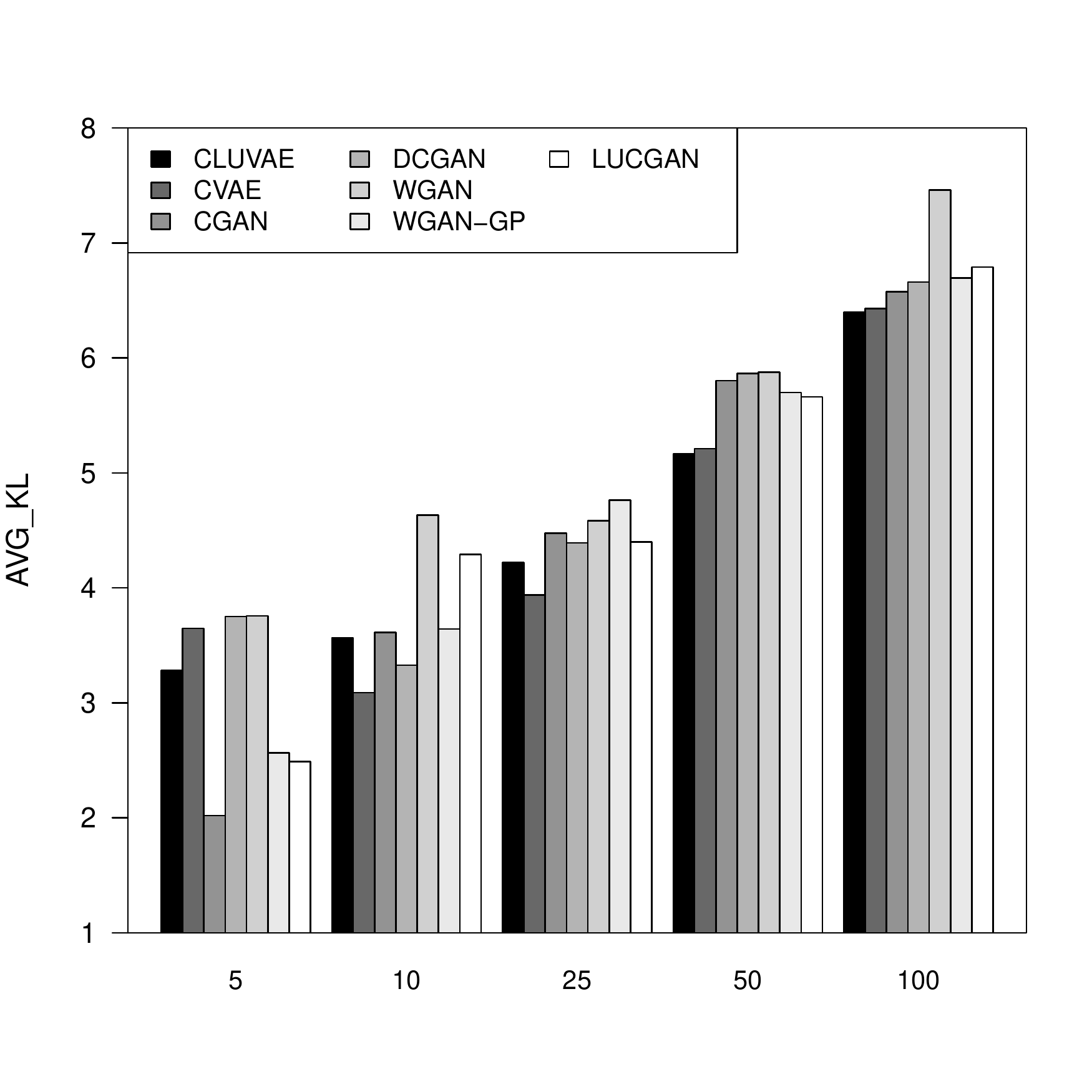} \label{fig:ss_avg_kl}}
	\subfigure[AVG\_JS]{\includegraphics[width=4.35cm]{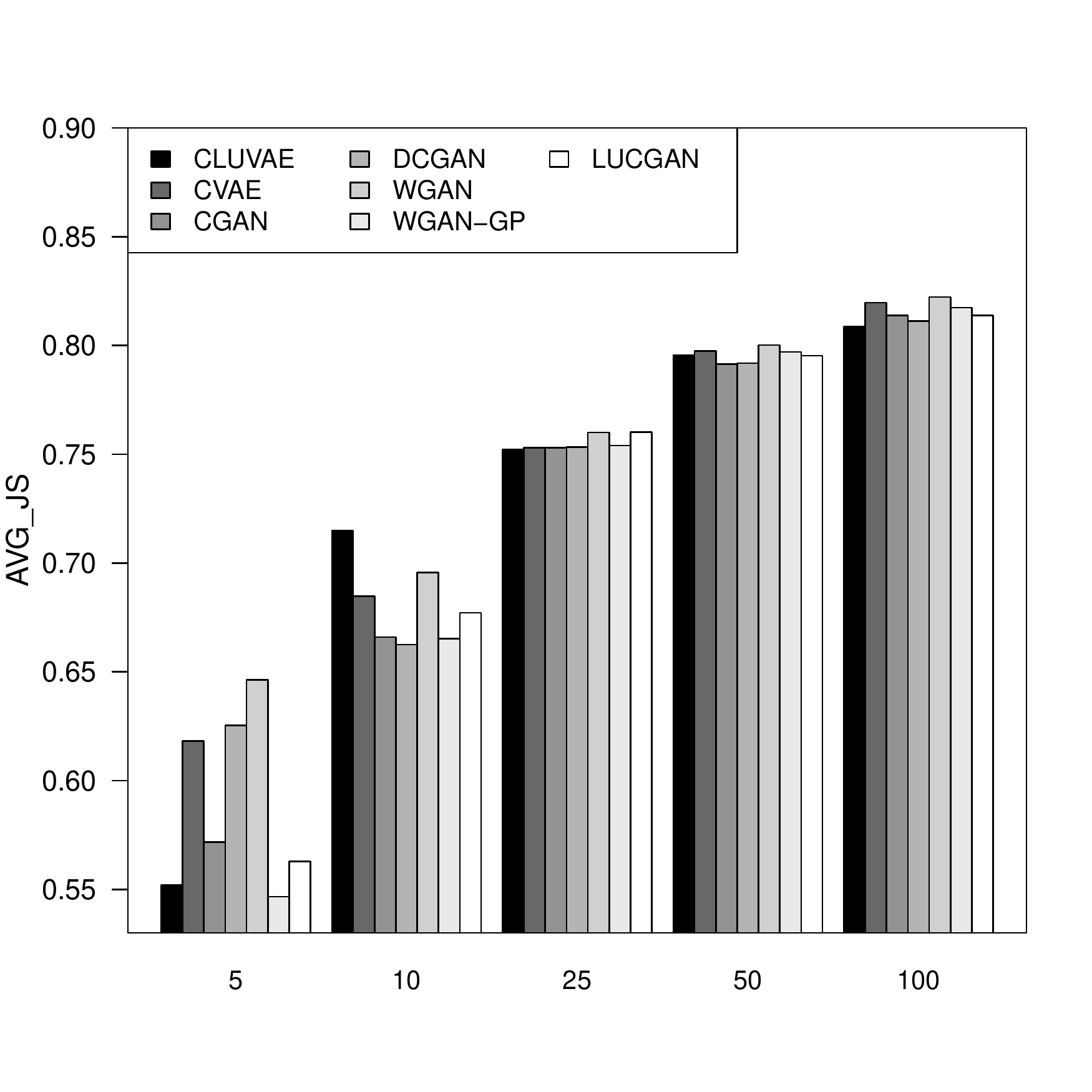}\label{fig:ss_avg_js}}
	\subfigure[AVG\_HD]{\includegraphics[width=4.35cm]{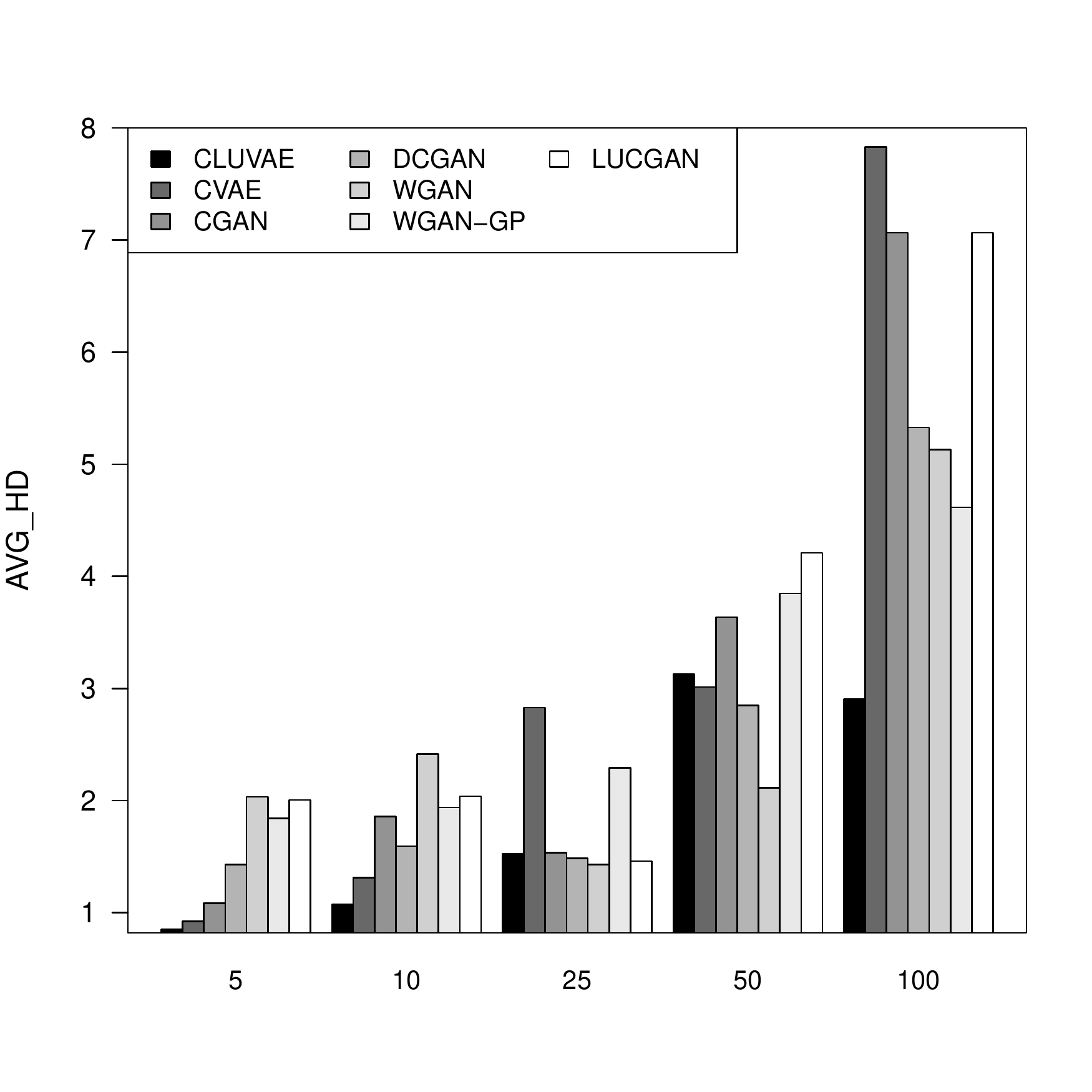}\label{fig:ss_avg_hd}}
	\subfigure[AVG\_Cos]{\includegraphics[width=4.35cm]{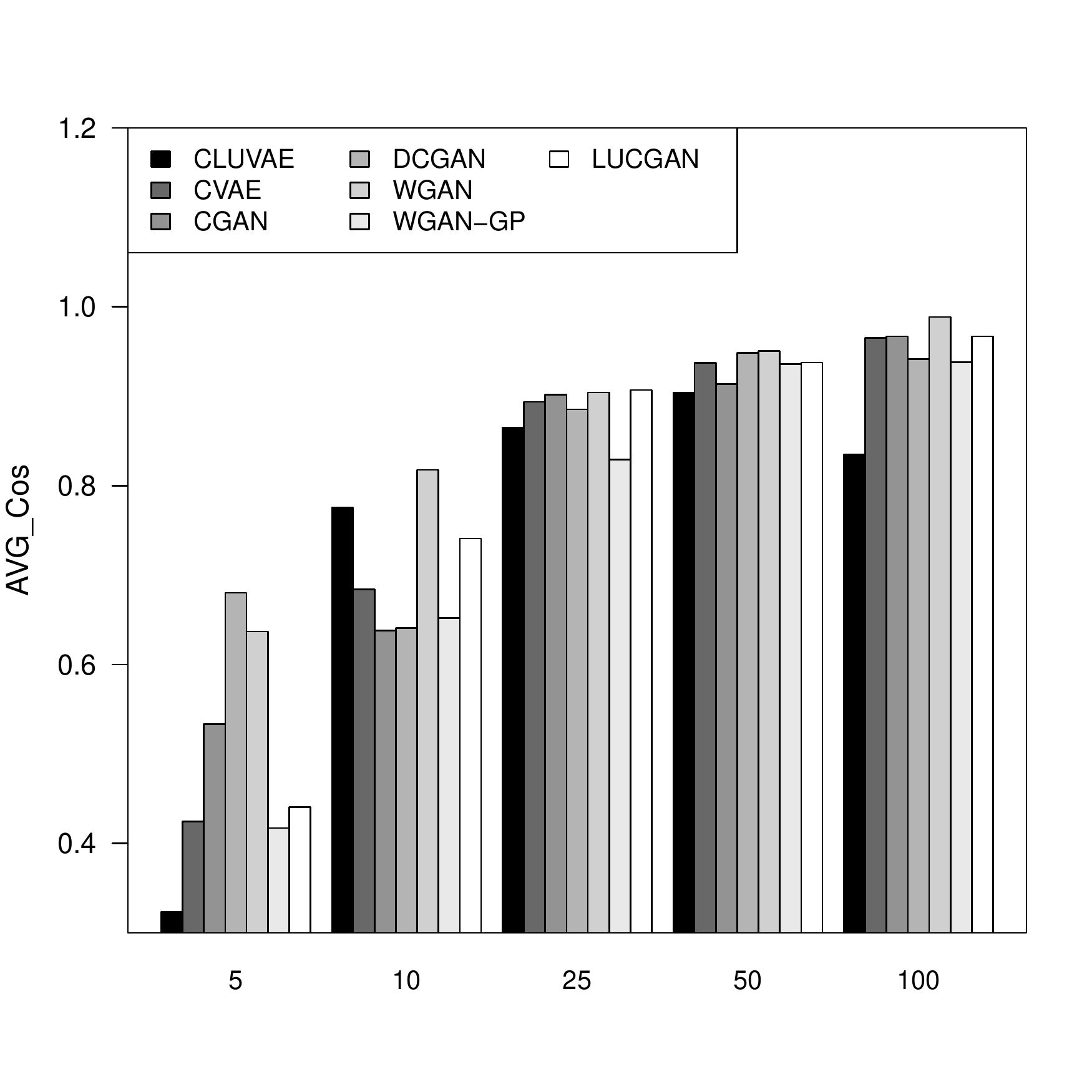}\label{fig:ss_avg_Cos}}	
	\vspace{-0.25cm}
	\caption{The influence of different square sizes for land-use configuration generation.}
	\label{fig:square_size}
	\vspace{-0.4cm}
\end{figure*}

\subsection{Ablation study for CLUVAE \textbf{Q3}}
We conducted an ablation study of CLUVAE to validate the necessity of each part of CLUVAE.
Figure \ref{fig:ablation_study} shows the performances of CLUVAE, CLUVAE*, CLUVAE', and CLUVAE- in terms of all evaluation metrics.
We can find that CLUVAE outperforms other variants with great improvements, which indicates the effectiveness of the complete CLUVAE.

In addition, another interesting observation that the performance of CLUVAE* is worse than the performance of CLUVAE'.
From another perspective, this observation reflects that in the land-use configuration generation task, human instructions are more important than the surrounding contexts.
This conclusion is reasonable because the surrounding contexts' features only provide environmental constraints to the generative model, but human instructions determine that the generated land-use configuration belongs to which green level.

Moreover, a careful inspection of Figure \ref{fig:ablation_study} shows that CLUVAE- has a big gap to CLUVAE in terms of all metrics.
The observation reveals that the hierarchical relationship between urban functional zones and land-use configurations plays an important role in the configuration generation task.
In summary, this experiment validates that each part of CLUVAE is indispensable and necessary.

\subsection{The study of the influence of square sizes \textbf{Q4}}
To study the influence of the square size on configuration generation, we set $N=5$, $N=10$, $N=25$, $N=50$, $N=100$ respectively to compare the generative performance.
Here, the smaller value of $N$ is, the larger size of square is.
Figure \ref{fig:square_size} shows the comparison results.
We can find that with the increase of the square size, the value of all metrics decreases. 
A possible explanation for the observation is that the larger square size makes the information of land-use configuration contain more coarse-grained and simpler.
The generative model can capture the characteristics of the kind of configurations more easily, thus, the model performance becomes better. 

In addition, another interesting observation
is that CLUVAE beats other baseline models in terms of all metrics no matter the square size is.
Meanwhile, a potential interpretation for the observation is that the learned distribution by CLUVAE really captures the characteristics of land-use configurations belongs to different green levels.
In summary, the results show the robustness and superiority of CLUVAE from a square size perspective compared with other baseline models.

\begin{figure*}[htbp]
\centering
\subfigure[Original land-use configuration under different human guidance]{
\begin{minipage}[t]{1.0\linewidth}
\centering
\includegraphics[width=7in]{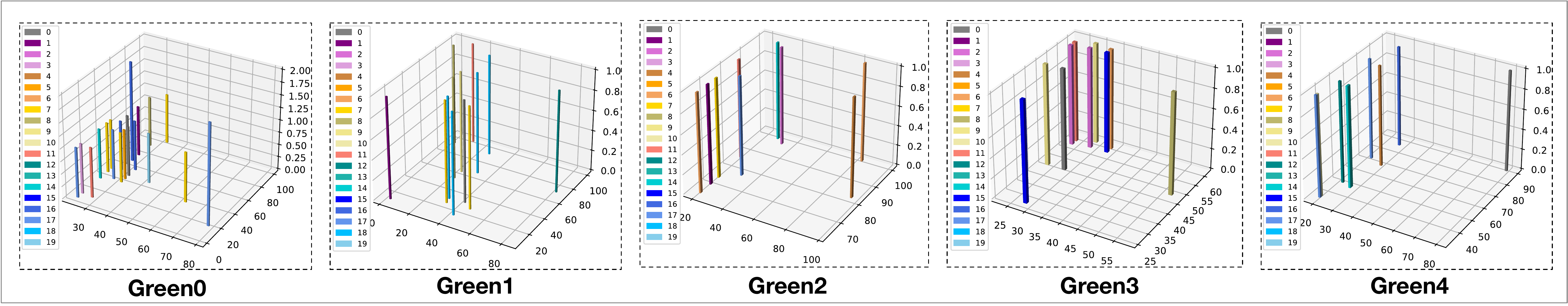}
\label{fig:original_land}
\end{minipage}%
}%
\vspace{-0.1cm}

\subfigure[Generated land-use configuration under different human guidance]{
\begin{minipage}[t]{1.0\linewidth}
\centering
\includegraphics[width=7in]{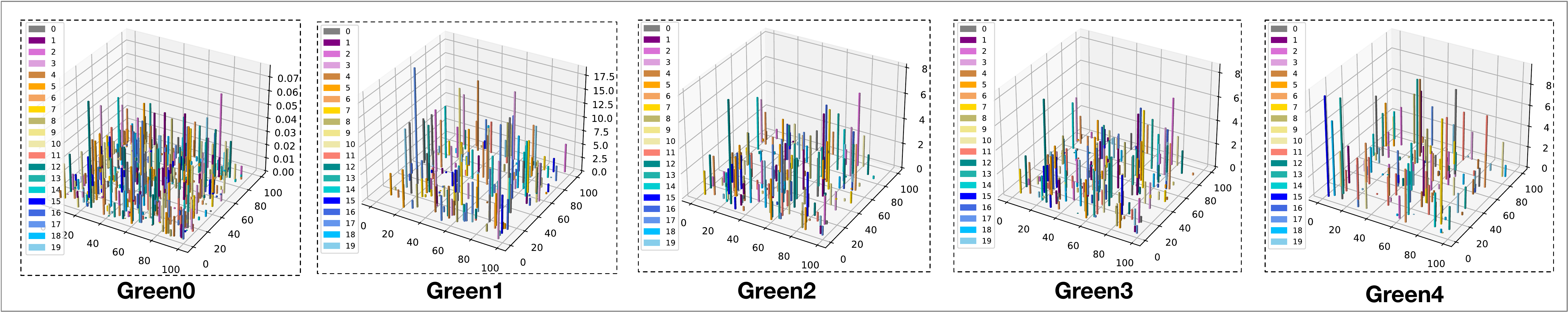}
\label{fig:generate_land}
\end{minipage}
}%
\centering
\vspace{-0.3cm}
\caption{Visualization comparison between original land-use configurations and generated land-use configurations }
\label{fig:visual_land}
\vspace{-0.6cm}
\end{figure*}

\subsection{Comparison study between original and generated land-use configurations under different human guidance \textbf{Q5}}

To understand the function of human instructions in land-use configuration generation and observe the visual differences between the original and generated land-use configurations,
we selected the most representative original and generated land-use configurations under different human instructions, and displayed these configurations in a 3-dimensional space.
Figure \ref{fig:visual_land} shows the visualization and comparison results, in which the left color legend indicates the mapping relations between the number of POI categories and colors;
the right 3D part reflects the POI distribution of land-use configurations;
the height of each bar indicates the number of POIs at the corresponding position;
the label of each subfigure is the corresponding human instructions (\textit{i.e., green level label}).
A careful inspection for Figure \ref{fig:original_land} and Figure \ref{fig:generate_land} shows that the generated configurations are organized and contain enough planning information for implementation in realistic.
A possible explanation for the observation is that CLUVAE captures the pros of land-use configurations under different green levels.
Thus, the CLUVAE brings these characteristics into land-use configuration generation.

In addition, for the green level of land-use configurations, Green0 $<$ Green1 $<$Green2 $<$ Green3 $<$ Green4.
Through comparing the inner subfigures in Figure \ref{fig:original_land} and Figure \ref{fig:generate_land}, we can find that with the increase of the green level, the POI distribution of land-use configurations becomes more sparse.
A potential interpretation for the observation is that sparse POI distribution indicates that the corresponding land-use configuration has empty geographical spaces.
Thus, planners can utilize these spaces to plant green plants for improving the green level of the land-use configuration.
In summary, this experiment shows that CLUVAE is capable of capturing the characteristics of land-use configurations belonging to different green levels.
Thus, urban planners can use CLUVAE to customize land-use configurations based on their requirements.

\section{Related Work}

\textbf{Data Fusion}
refers to integrating multiple data sources for capturing more accurate and valuable information ~\cite{meng2020survey}.
In reality, there are many application scenarios of artificial intelligence that need the data fusion process to capture more informative features for better modeling. 
For instance, Wang {\it et al.} ~\cite{wang2021reinforced} proposed a user profiling framework based on reinforcement learning via analyzing the check-in data and geographical POI data.
Liu {\it et al.} ~\cite{liu2018modeling} modeled the interaction coupling of multi-view spatiotemporal contexts to predict shared-bike destination. 
Wang {\it et al.} ~\cite{wang2021measuring} studied the urban vibrancy of residential communities via analyzing the multiple data resources such as check-ins, urban geography data, and human mobility.
Wang {\it et al.} ~\cite{wang2020defending} fused the monitoring value of different kinds of sensors together to detect the anomaly status of water treatment plants.
Compared with these works, in this paper, we face more complicated data resources.
Considering the complexity of urban planning, we employ a spatial attributed graph to organize all socioeconomic characteristics of surrounding contexts together, and utilize a graph embedding method to preserve these characteristics into an embedding vector.
Moreover, for integrating the information of human guidance and surrounding context, we concatenate the one-hot vector of human guidance and the embedding of surrounding contexts as our model input.

\textbf{Deep Variational Auto Encoder (VAE)} is a classical kind of deep generative model ~\cite{oussidi2018deep}.
Traditional VAE models achieve great success in many generated applications such as image generation, text generation, etc ~\cite{zhu2020s3vae,luo2020mg,akbari2018semi}.
For instance, 
Zhang {\it et al.} ~\cite{zhang2021image} proposed a framework, namely VAE-AGAN, which integrates VAE and generative network (GAN) together to augment regions of interest (ROIs) image data for improving the ROIs detection of the model.
Chen {\it et al.} ~\cite{chen2021trajvae} utilized a VAE model to generate trajectories based on the characteristic of trajectories learned by LSTM for fulfilling the needs of self-driving simulation and other traffic analysis tasks.
However, although VAE can generate data samples by simulating the data distribution, the output of VAE cannot be controlled.
Conditional VAE is proposed to resolve the limitation, and obtain remarkable achievements ~\cite{jiang2020transformer,zhang2020disentangling}.
For example, Yonekura {\it et al.} \cite{yonekura2021data} developed a new turbine designer
based on the conditional VAE technique for improving the design efficiency.
Compared with these works, our CLUVAE is also based on the conditional VAE setting.
But we add a regularization item to the loss function during the optimization process for producing more reasonable results.

\textbf{Urban Planning} is vital for the future development of a geographical area, which designing the land use and environment of the area according to human requirements and surrounding contexts ~\cite{oliveira2010evaluation,naess2001urban}.
However, to attain effective urban planning solutions, human experts have to spend much attention to consider lots of constraints such as government policy, environmental protection, etc.
For instance, Ratcliffe {\it et al.} ~\cite{ratcliffe2009urban} studied the relationship between urban planning and real estate development.
Papa {\it et al.} ~\cite{papa2016smart} studied how to build up a sustainable smart city via urban planning.
Recently, with the development of deep learning, many researchers bring deep models into the urban planning domain ~\cite{wang2020reimagining,nguyen2020simulating}.
For instance, Wang {\it et al.} ~\cite{wang2020reimagining} utilized GAN-based framework to generate land-use configuration based on surrounding contexts.
Compared with these works, CLUVAE has higher customization ability and generates more reasonable configurations.

\section{Conclusion Remarks}

In this paper, we propose a deep land-use configuration generation framework, namely CLUVAE, which can generate land-use configurations according to human guidance and considering surrounding contexts' socioeconomic features.
To produce personalized and reasonable land-use configuration, and make our generation model robust, we formulate the automated urban planning process
into a deep variational autoencoder framework.
We implement the framework based on an encoder-decoder paradigm.
Specifically, we first utilize the encoder part to learn a distribution based on  condition embeddings (\textit{i.e., surrounding contexts' feature embedding + human guidance embedding}) and land-use configurations.
Then, a variational Gaussian embedding mechanism is used to produce latent embeddings by sampling from the learned distribution.
After that, we input condition embeddings and latent embeddings into the decoder part for reconstructing land-use configurations and the corresponding urban functional zones.
Finally, the well-trained decoder is our expected land-use configuration generator.
Extensive experiments validate the superiority and effectiveness of CLUVAE.
Although CLUVAE could obtain desirable results, it is also subject to certain limitations.
In the future, we will focus on utilizing sophisticated human guidance to improve generative performance.
Because complex human guidance can provide more information for generating reasonable land-use configurations.
 Thus, how to deal with sophisticated human guidance in planning generation is our next focus.


\section*{Acknowledgment}
This research was partially supported by the National Science Foundation (NSF) via the grant numbers: 1755946, 2040950, 2006889, 2045567.


\bibliographystyle{IEEEtran}    
\bibliography{ref}

\end{document}